\documentclass[10pt,twocolumn,letterpaper]{article}

\usepackage{cvpr}              
\usepackage{url}
\usepackage{graphicx}
\usepackage[accsupp]{axessibility}  
\usepackage[accsupp]{axessibility} %
\usepackage[table]{xcolor}   
\usepackage{booktabs}        
\usepackage{array}           
\usepackage{tabularx}        
\usepackage{multirow}        
\usepackage{colortbl}        

\usepackage{caption}         
\usepackage{subcaption}      
\usepackage{amssymb}
\usepackage{natbib}          
\usepackage{csquotes}        
\usepackage{enumitem}        

\definecolor{tabfirst}{rgb}{1,0.7,0.7}   
\definecolor{tabsecond}{rgb}{1,0.85,0.7} 

\newcolumntype{C}{>{\centering\arraybackslash}X} 
\renewcommand{\arraystretch}{1.4}               
\usepackage{pifont}       
\usepackage{amsmath}      
\definecolor{cvprblue}{rgb}{0.21,0.49,0.74}
\usepackage[pagebackref,breaklinks,colorlinks,allcolors=cvprblue]{hyperref}

\def\paperID{***} 
\def\confName{CVPR}
\def\confYear{2026}

\title{ExtrinSplat: Decoupling Geometry and Semantics for Open-Vocabulary Understanding in 3D Gaussian Splatting}

\author{
	Jiayu Ding\textsuperscript{1,*}, 
	Xinpeng Liu\textsuperscript{1,*}, 
	Zhiyi Pan\textsuperscript{2}, 
	Shiqiang Long\textsuperscript{3}, 
	Ge Li\textsuperscript{1,\dag}
	\vspace{-6pt} 
	\\ \vspace{-8pt} 
	{\footnotesize \textsuperscript{1}Guangdong Provincial Key Laboratory of Ultra High Definition Immersive Media Technology, Shenzhen Graduate School, Peking University} \\ \vspace{-8pt}
	{\footnotesize \textsuperscript{2}School of Computer Science and Technology, Tianjin University} 
	{\footnotesize \textsuperscript{3}Guangdong Bohua UHD Innovation Center Co., Ltd.} \\ \vspace{-8pt}
	{\footnotesize \texttt{\{jyding25@stu, lxpeng@stu, geli@ece\}.pku.edu.cn, zhypan@tju.edu.cn, longsq@bohuauhd.com}} \\ \vspace{-4pt}
	{\footnotesize \textsuperscript{*}Equal contribution. \quad \textsuperscript{\dag}Corresponding author.}
}

\begin{document}
\maketitle
\begin{abstract}
	Lifting 2D open-vocabulary understanding into 3D Gaussian Splatting (3DGS) scenes is a critical challenge. Mainstream methods, built on an embedding paradigm, suffer from three key flaws: (i) geometry-semantic inconsistency, where points, rather than objects, serve as the semantic basis, limiting semantic fidelity; (ii) semantic bloat from injecting gigabytes of feature data into the geometry; and (iii) semantic rigidity, as one feature per Gaussian struggles to capture rich polysemy.
	To overcome these limitations, we introduce ExtrinSplat, a framework built on the extrinsic paradigm that decouples geometry from semantics. Instead of embedding features, ExtrinSplat clusters Gaussians into multi-granularity, overlapping 3D object groups. A Vision-Language Model (VLM) then interprets these groups to generate lightweight textual hypotheses, creating an extrinsic index layer that natively supports complex polysemy.
	By replacing costly feature embedding with lightweight indices, ExtrinSplat reduces scene adaptation time from hours to minutes and lowers storage overhead by several orders of magnitude. On benchmark tasks for open-vocabulary 3D object selection and semantic segmentation, ExtrinSplat outperforms established embedding-based frameworks, validating the efficacy and efficiency of the proposed extrinsic paradigm.
\end{abstract}    
\section{Introduction}
\label{sec:intro}
Open-vocabulary 3D scene understanding enables the parsing of 3D scenes with arbitrary natural language queries, moving beyond the limitations of predefined categories to offer enhanced generalization and richer semantics for applications like autonomous driving~\citep{kong2025multi,zhu2023understanding} and robotics~\citep{chen2024sugar,song2025robospatial}. The primary challenge in this domain lies in finding an efficient and effective 3D scene representation. Traditional methods such as voxels, point clouds, and meshes, while useful for structure modeling, struggle with the trade-off between detail and computational expense. 
Recently, 3D Gaussian Splatting (3DGS)~\citep{kerbl_3d_2023} has been proposed, which achieves high-fidelity modeling and rendering while maintaining high rendering speeds, making it an ideal foundation for next-generation 3D scene understanding.

Recently, some methods have leveraged 3DGS to achieve point-level open-vocabulary 3D scene understanding. A majority of these methods~\citep{li_instancegaussian_2025,wu2024opengaussian,liang2024supergseg,sun2025cags,yin2025semantic} attempt to embed high-dimensional semantic features into each 3D Gaussian point, optimizing these features via contrastive learning. This requires tens of thousands of mask-guided contrastive learning iterations, incurring substantial optimization costs. More recently, a few approaches~\citep{jun-seong_dr_2025,jiang2025votesplat,marrie2025ludvig} have explored embedding semantic features into each Gaussian via matching, which optimizes costs and achieves strong results. However, all these methods are built upon a common \textbf{embedding paradigm} that intrinsically embeds semantic features into the 3D Gaussian points. While this paradigm has shown promising results, it suffers from three limitations:
\textbf{1)~Geometry-Semantic Inconsistency:} Objects, not Gaussian points, should be the basic unit of semantic understanding. Gaussians are designed to express scene geometry, not specifically for semantics. 
Moreover, this inherent disparity between geometric and semantic representations manifests at object boundaries as what we term \enquote{neutral points}: points driven solely by geometric optimization for high-fidelity boundary rendering, which intrinsically lack a semantic assignment.
Embedding paradigm, by forcibly assigning semantic features to every Gaussian, inevitably lead to semantic ambiguity and confusion at object boundaries.
\textbf{2)~Semantic Bloat:} The core of embedding paradigm is to lift and store 2D visual features. This results in 3DGS scenes being injected with Gigabytes (GB) of feature data, dramatically increasing storage and downstream processing burdens.
\textbf{3)~Semantic Rigidity:} A single Gaussian can only store one visual feature vector. In reality, a single Gaussian point may be part of multiple objects, thereby possessing distinct semantic meanings that differ significantly in the feature space. For example, a Gaussian point on a car’s window surface can be identified as \enquote{window}, but it is also correctly described as part of the \enquote{car} itself.
The existing embedding paradigm attempts to forcefully fuse a point's multiple, and often conflicting, semantic identities into one feature vector via contrastive learning or feature projection. 
This compromised representation not only causes inherent semantic inaccuracy but also fundamentally limits the model's ability to express the rich polysemy of complex scenes.

To overcome these limitations, we propose the \textbf{extrinsic paradigm}, a distinct, decoupled and layered architecture. This paradigm avoids injecting semantic features into the Gaussian points. Instead, it models semantics as an independent, abstract index layer. This semantic layer references the underlying geometry, leaving its original structure intact. The core advantage of this decoupled design is that it operates on the natural atomic units of each domain: Gaussian points for geometry and objects for semantics.
Based on this extrinsic paradigm, we propose ExtrinSplat, a point-level open-vocabulary 3D understanding framework. ExtrinSplat addresses the three inherent drawbacks of the embedding paradigm:
\textbf{1)~Addressing Geometry-Semantic Inconsistency}: ExtrinSplat shifts the semantic unit from points to objects by clustering Gaussian points into distinct, per-entity 3D groups via multi-view mask back-projection. A dedicated mechanism further identifies and excludes neutral points from semantic assignment.
This enhances semantic clarity at object boundaries.
\textbf{2)~Addressing Semantic Bloat}: We avoid storing high-dimensional feature vectors. Instead, we use Vision-Language Models (VLMs) to directly interpret these 3D object groups and generate candidate \enquote{textual hypotheses} for each group. Semantics are then stored as lightweight extrinsic indices pointing to these hypotheses, reducing storage overhead by several orders of magnitude compared to feature embedding.
\textbf{3)~Addressing Semantic Rigidity}: We introduce a multi-granularity, overlapping object grouping mechanism. The same Gaussian point can simultaneously belong to multiple groups with different semantic identities. Each group links to multiple textual hypotheses, natively supporting the rich polysemy unattainable by the embedding paradigm.

Our contributions are summarized as follows:
\begin{itemize}
	\item We propose ExtrinSplat, a new framework realizing the extrinsic paradigm, which efficiently decouples 3D geometry and semantics through object grouping and lightweight textual indices.
	\item We introduce a multi-granularity, overlapping object grouping strategy, enabling the framework to natively support rich semantic polysemy.
	\item We define the concept of neutral points and propose a dedicated handling mechanism to address this issue.
\end{itemize}

\section{Related Works}
\subsection{Preliminary: 3D Gaussian Splatting} 
3D Gaussian Splatting (3DGS)~\citep{kerbl_3d_2023} models 3D scenes with explicit 3D Gaussians, enabling high-quality, real-time rendering. It represents a scene as a collection of 3D Gaussians $\mathcal{G} = \{g_i\}_{i=1}^N$, each defined by its position, covariance (governing scale and orientation), color, and opacity. To generate a 2D image, these 3D Gaussians are projected onto an image plane and then blended in a depth-sorted order via \enquote{splatting}. The final color $C(p)$ for any pixel $p$ is determined through alpha compositing~\citep{munkberg2022extracting}:
\begin{equation}
	\label{eqw}
	C(p) = \sum_{i=1}^{|\mathcal{G}_p|} c_{g_i^p} \alpha_{g_i^p} \prod_{j=1}^{i-1} (1 - \alpha_{g_j^p}),
\end{equation}
where $c_{g_i^p}$ and $\alpha_{g_i^p}$ are the color and opacity of the $i$-th Gaussian in the sorted set for pixel $p$. The product term, $\prod_{j=1}^{i-1} (1 - \alpha_{g_j^p})$, calculates the accumulated transmittance, which represents the light that reaches the $i$-th Gaussian after passing through all prior ones.

\subsection{Open-Vocabulary Understanding in 3DGS}
Prevailing methods for semanticizing 3DGS for open-vocabulary understanding follow two primary paradigms: pixel-based and point-based. 
Pixel-based methods employ a \enquote{render-then-match} paradigm: they first render the entire scene into dense 2D feature maps and subsequently perform semantic matching in the image space. In contrast, point-based methods adopt a \enquote{match-then-render} strategy, first identifying a sparse set of semantically relevant 3D points and then rendering only this pre-filtered subset.

In pixel-based methods, Feature-3DGS~\citep{zhou_feature_2024} distills semantic features from 2D foundation models into 3DGS, enabling fast semantic rendering. LEGaussians~\citep{shi_language_2024} adds uncertainty and semantic to each Gaussian and compares rendered semantic maps with quantized CLIP~\citep{radford_learning_2021} and DINO~\citep{zhang_dino_2022} features. LangSplat~\citep{qin_langsplat_2024} learns language features in a scene-specific latent space and renders them as semantic maps. GS-Grouping~\citep{ye_gaussian_2024} assigns a compact identity encoding to each Gaussian and leverages masks from the Segment Anything Model (SAM)~\citep{kirillov_segment_2023} for supervision. GOI~\citep{qu_goi_2024} introduces an optimizable semantic hyperplane to separate pixels relevant to language queries, improving open-vocabulary accuracy.
\begin{figure*}[t!]
	\centering
	\includegraphics[width=1\linewidth]{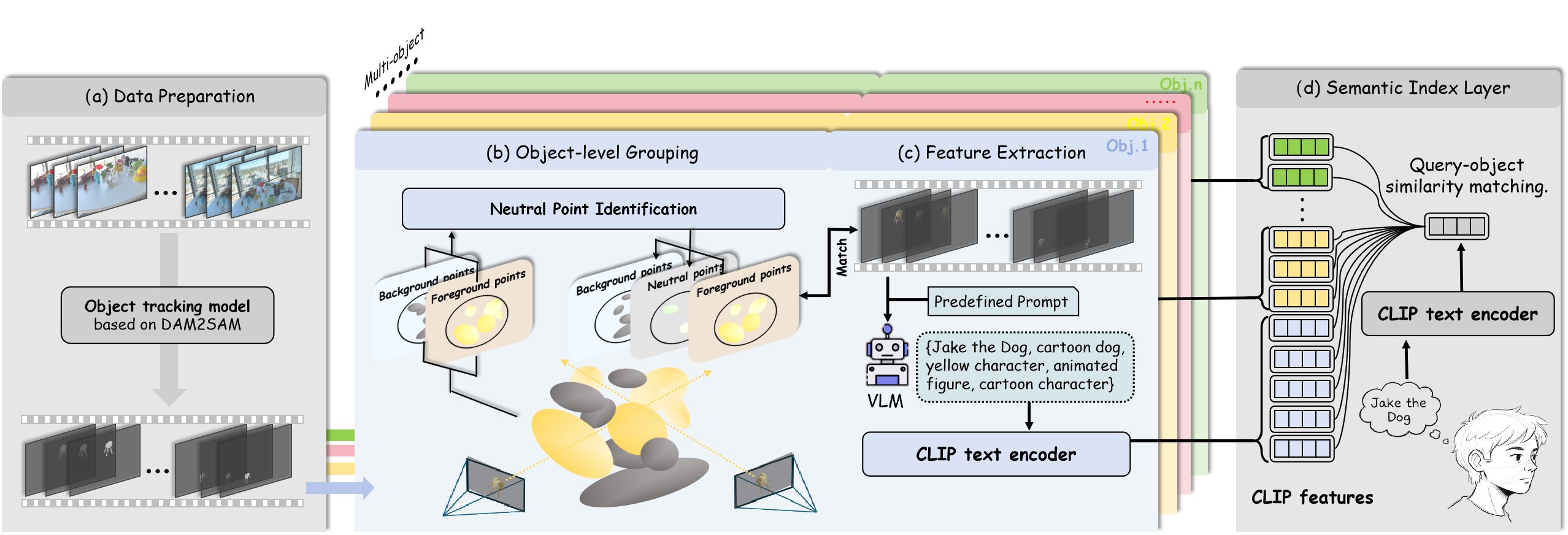}
	\caption{Overview of our method. (a) Multi-view 2D segmentation masks are first extracted from the input scene. (b) Based on these masks, our method lifts the objects into 3D point groups via back-projection, refining their boundaries by filtering ambiguous neutral points. (c) Each refined group is then grounded by using a VLM to generate textual hypotheses from its key views, which are encoded into semantic features via a CLIP text encoder. (d) Finally, these geometric groups and semantic features are assembled into an extrinsic semantic index layer, enabling open-vocabulary querying by matching a user's text query against the pre-computed features. }
	\label{014}
\end{figure*}

However, these methods rely on rendered 2D semantic maps, so reasoning remains in 2D. They lack awareness of 3D structure and are thus less suited for tasks requiring direct 3D interaction, such as embodied intelligence. To overcome these limitations, point-based methods have been proposed. These approaches operate on a foundational \enquote{select-then-render} pipeline. OpenGaussian~\citep{wu2024opengaussian} uses SAM masks to learn instance features with 3D consistency, introduces a two-stage codebook for feature discretization, and links 3D points with 2D masks and CLIP features for open-vocabulary selection. InstanceGaussian~\citep{li_instancegaussian_2025}, based on Scaffold-GS~\citep{lu_scaffold-gs_2024}, jointly learns appearance and semantics and adaptively aggregates instances, reducing semantic–appearance misalignment. Dr.Splat~\citep{jun-seong_dr_2025} registers 2D CLIP features to 3D Gaussian points via aggregation and compresses them using a pre-trained Product Quantization codebook. LUDVIG~\citep{marrie2025ludvig}  uplifts 2D CLIP features to 3D Gaussian points via weighted aggregation and refines them using a DINOv2-based~\citep{oquab2023dinov2} graph diffusion mechanism.
All these approaches adhere to an embedding paradigm, intrinsically binding high-dimensional features to the Gaussian geometry.

\section{Method}

\subsection{Overall Architecture}
\label{sec:overall_architecture}
We present ExtrinSplat, a training-free framework that realizes the extrinsic paradigm by decoupling 3D geometry from semantics, as shown in Fig.~\ref{014}.
The core of the ExtrinSplat framework is to treat each object in a 3D scene as an independent entity.  
For every object, the model performs separate 3D grouping and semantic feature assignment. The assignment process for these objects is mutually independent and can match the same Gaussian points simultaneously.
User queries in natural language are matched directly against the features of these independent object groups. This approach eliminates the need to store a unique semantic feature vector for every Gaussian point, enabling more efficient and coherent object-level semantic understanding.

Our method takes an optimized 3DGS scene representation and its corresponding image sequence as input.
First, the data preparation stage (\S\ref{sec:data_prep}) generates multi-view, multi-granularity object masks.
Next, the object-level grouping stage (\S\ref{sec:grouping_main}) links 2D masks to 3D Gaussian points and purifies the resulting object boundaries by identifying and excluding ambiguous neutral points.
Then, the instance feature extraction stage (\S\ref{sec:feature_extraction}) uses a VLM to generate textual hypotheses for each object group.
Finally, these geometric groups and semantic components are assembled into the extrinsic semantic index layer (\S\ref{sec:index_layer}) to formalize the decoupling and enable efficient open-vocabulary querying.

\subsection{Data Preparation}
\label{sec:data_prep}
Our goal is to extract a comprehensive set of multi-view segmentation masks for all objects in a scene. To this end, we first employ SAM on the initial frame, $I_0$, leveraging its ability to produce object masks at three distinct granularity levels (e.g., part, object, scene). To ensure stable tracking throughout the sequence, especially in complex scenes with visually similar distractors, we employ the DAM2SAM~\citep{videnovic2025distractor} model. Its specialized distractor-aware memory is crucial for maintaining accurate object identities where other trackers might fail. To capture new objects that appear later, we introduce a periodic detection mechanism that re-segments the scene at fixed intervals and identifies new instances based on a minimal IoU overlap criterion with existing tracks. The entire pipeline, from tracking to new object detection, is executed independently for each of the three granularity levels to yield a complete and hierarchical set of masks.

Our pipeline is designed for robustness, as potential data preparation artifacts, such as tracking failures or re-identification errors, are gracefully handled by our downstream object grouping and query matching modules. This design minimizes the requirements for perfect input data (see Appendix for details). This independent, multi-granularity processing ensures the intrinsic overlap (e.g., fine-grained part within coarse-grained object) of the mask set, providing a foundation for the model's subsequent polysemous understanding.

\subsection{Object-level Grouping}
\label{sec:grouping_main}
To realize multi-granularity grouping, we apply our grouping strategy independently to each object mask set generated in Sec.~\ref{sec:data_prep}. This parallel and independent execution enables a single 3D Gaussian to be assigned to multiple object groups (e.g., \enquote{window} and \enquote{car}), thereby natively accommodating semantic polysemy. Specifically, for each group, we first identify the object's high-confidence core via mask back-projection, then refine its boundaries by identifying and excluding ambiguous points with our neutral point processing module, ensuring a clean result.

\noindent \textbf{Initial 3D Grouping via Mask Back-projection.} We link 2D masks to 3D Gaussian points by back-projecting them to estimate a per-point foreground probability. For each object, we process its multi-view masks, first discarding any null (entirely black) masks from viewpoints where the object is unseen. For each valid mask, we then cast a ray through each pixel $r$ and sum the contributions of all intersected Gaussians. The contribution of the $j$-th Gaussian $G_j$ along ray $r$ is determined by its accumulated transmittance and opacity, defined as:
\begin{equation}
	w(r, G_j) = T(r, G_j) \cdot \alpha(r, G_j),
\end{equation}
where $T(r, G_j)$ denotes the accumulated transmittance up to $G_j$, and $\alpha(r, G_j)$ is its effective opacity. To ensure design consistency, we define $w(r, G_j)$, representing the contribution of Gaussian $G_j$ to pixel $r$, to be identical to the forward color rendering weight of 3DGS given in Eq.~\ref{eqw}.

For each 3D Gaussian point $G_j$, we compute its total foreground ($W_1$) and background ($W_0$) weights, corresponding to $k=1$ and $k=0$ respectively, by aggregating contributions from multi-view 2D masks:
\begin{equation}
	W_k(G_j) = \sum_{v \in \mathcal{V}} \sum_{r \in \mathcal{P}_v} \delta(m_v(r) - k) \cdot w_v(r, G_j),
\end{equation}
where $\mathcal{V}$ is the set of visible views, $\mathcal{P}_v$ the pixels in a view, $m_v(r)$ the mask value, $\delta(\cdot)$ the indicator function, and $w_v(r, G_j)$ the contribution weight.
Based on these weights, we form an initial foreground set, $\mathcal{F}$, using a simple hard assignment: $\mathcal{F} = \{ G_j \mid W_1(G_j) > W_0(G_j) \}$. All remaining points are consequently assigned to the background. 

\noindent \textbf{Neutral Point Processing.} 
During rendering, it is inevitable that some points lie at the boundaries between objects but do not semantically belong to any specific category. We refer to these points as neutral points. Their semantic assignment directly affects the accuracy of rendered object boundaries. Existing methods typically assume that each 3D Gaussian point belongs either to the foreground or to the background, i.e., every point has a clear semantic label. In practice, however, many points at boundaries are transitional and may not carry a well-defined semantic meaning. Such points should be considered neither foreground nor background. Our goal is to identify and exclude these neutral points from semantic supervision, thereby mitigating potential artifacts and improving the accuracy of the final segmentation.

To identify neutral points, we leverage multi-view semantic consistency. While points deep within an object are consistently labeled across views, those near boundaries often exhibit conflicting semantics. To quantify this ambiguity, we treat each viewpoint as providing a discrete semantic label for a given Gaussian point. Specifically, for each point $p$, we project its center into every visible view and record whether it lands inside (foreground) or outside (background) the corresponding 2D mask. This process yields a set of binary labels $\{l_v\}_{v \in \mathcal{V}}$ for each 3D point. The semantic entropy $H(p)$, which quantifies the disagreement among these discrete labels, is calculated as:
\begin{equation}
	H(p) = - \left( \frac{V_f}{V} \log_2 \frac{V_f}{V} + \frac{V_b}{V} \log_2 \frac{V_b}{V} \right),
\end{equation}
where $V_f$ and $V_b$ are the respective counts of foreground and background labels within the set $\{l_v\}_{v \in \mathcal{V}}$, and $V = |\mathcal{V}| = V_f + V_b$. Points with entropy $H(p)$ exceeding a threshold $\tau_h$ form an initial candidate set $\mathcal{C}$ of ambiguous points.

This set $\mathcal{C}$ is impure, containing both true neutral points used for smooth blending and mislabeled solid points that belong to an object's surface. To distinguish them, we use a geometric property: opacity ($\alpha$). Points on solid surfaces typically have high opacity, while transitional points used for anti-aliasing have low opacity. We filter $\mathcal{C}$ based on this idea: if a point $p \in \mathcal{C}$ has an opacity $\alpha(p) > \tau_\alpha$, we classify it as a mislabeled solid point. These points, identified as part of a solid surface, are removed from the neutral candidate set $\mathcal{C}$, thereby retaining their initial classification as either foreground or background.

The remaining points in $\mathcal{C}$ are confirmed as the final neutral point set, which is excluded from all semantic supervision. The final set of foreground points is thus defined by the expression $\mathcal{F} \setminus \mathcal{C}$. Likewise, the background set is refined by removing these same points.
We use fixed values for the thresholds $\tau_h$ and $\tau_\alpha$ across all experiments for simplicity and robustness. A detailed sensitivity analysis on their selection is provided in Appendix.

\begin{figure}[t!]
	\centering
	\includegraphics[width=1\linewidth]{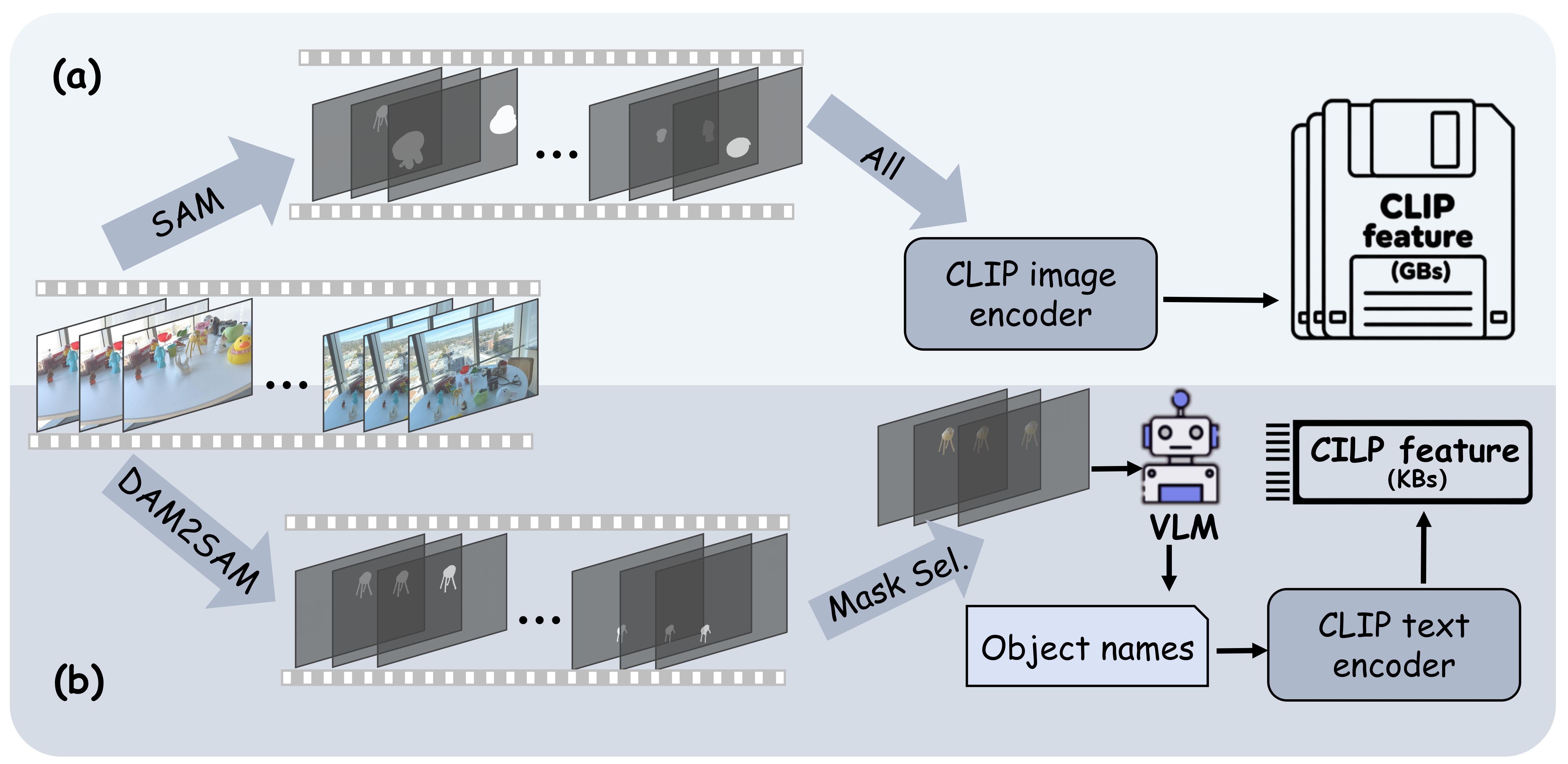}
	\caption{
		Comparison of 2D-3D feature association pipelines. 
		(a) Mainstream method (via direct extraction): All object masks, typically generated by SAM, are used to directly extract CLIP image features.
		(b) Our method (via semantic distillation): We leverage DAM2SAM to track a single instance. The top-N most visible masks are then interpreted by a VLM, distilling volatile visual appearances into a stable CLIP text representation derived from the generated object identity.
	}
	\label{002}
\end{figure}

\subsection{Instance Feature Extraction}
\label{sec:feature_extraction}
The dominant embedding paradigm, which intrinsically aggregates multi-view visual features into the 3D geometric structure, as shown in Fig.~\ref{002}(a), faces two key limitations:
\textbf{1)} Semantic instability. Aggregating visual features from multi-view 2D masks often produces biased and inconsistent 3D semantics. Due to viewpoint variance in 2D encoders (e.g., CLIP), the same 3D object can yield markedly different embeddings across views (see Appendix for visualization), reducing overall 3D accuracy.
\textbf{2)} Storage inefficiency. Extracting and storing high-dimensional visual features for every object mask in all views introduces heavy computational and storage costs, making the 3D scene representation inefficient and inflexible. 
To avoid these drawbacks, we propose semantic distillation: we use a VLM to interpret key views and generate textual hypotheses of the object's identity. As shown in Fig.~\ref{002}(b), this converts volatile visual appearances into a stable, canonical text representation.

Specifically, our semantic distillation process operates for each object group $i$ identified in Sec.~\ref{sec:grouping_main}. First, we select the top-$N$ masked views with the largest visible areas for this group. These views are fed into a Vision-Language Model (VLM) along with a predefined text prompt, which instructs the model to generate a set of candidate textual hypotheses (i.e., object names).  Our framework is VLM-agnostic; for this study, we employ Gemini 2.5 Pro~\citep{comanici2025gemini25pushingfrontier} , but other models can be readily substituted.
Next, these VLM-generated candidate names are encoded using a pre-trained CLIP text encoder. This process yields the group's final semantic component $\mathbf{Q}_i$, a set of feature vectors representing its identity, which is then passed to the extrinsic index layer described in the following section.

\subsection{Extrinsic Semantic Index Layer}
\label{sec:index_layer}

We construct the extrinsic semantic index layer to formalize the decoupling of geometry and semantics proposed by the extrinsic paradigm.
This structure is a set of object maps, $\mathcal{L} = \{ (\mathcal{G}_i, \mathbf{Q}_i) \}_{i=1}^N$, where $N$ is the total number of object groups.
Each map $i$ consists of a geometric component, $\mathcal{G}_i = \mathcal{F}_i \setminus \mathcal{C}_i$, which is the set of indices for all 3D Gaussian points in the group, and a semantic component, $\mathbf{Q}_i$, which is the set of pre-computed CLIP text features from the VLM's textual hypotheses.

Open-vocabulary querying becomes a fast lookup against this index layer. A text query is encoded into a feature vector $\mathbf{s}$ using the CLIP text encoder. This query $\mathbf{s}$ is then compared against the set of semantic features $\mathbf{Q}_i$ for each object group $i$. We use cosine similarity to quantify semantic relevance:
\begin{equation}
	\text{sim}(\mathbf{s}, \mathbf{q}) = \frac{\mathbf{s} \cdot \mathbf{q}}{\|\mathbf{s}\|\|\mathbf{q}\|}, \quad \mathbf{q} \in \mathbf{Q}_i.
\end{equation}
An object group $i$ matches if any of its semantic feature vectors $\mathbf{q} \in \mathbf{Q}_i$ exceeds a similarity threshold $\eta$. The set of matching group indices $\mathcal{I}_{\text{m}}$ is defined as:
\begin{equation}
	\mathcal{I}_{\text{m}} = \left\{ i \mid \max_{\mathbf{q} \in \mathbf{Q}_i} \text{sim}(\mathbf{s}, \mathbf{q}) > \eta \right\}.
\end{equation}
The final segmentation for the query is the union of the 3D Gaussian points from the matched geometric components:
\begin{equation}
	\mathcal{G}_{\text{final}} = \bigcup_{i \in \mathcal{I}_{\text{m}}} \mathcal{G}_i
\end{equation}

\begin{figure*}[!t]
	\centering
	\includegraphics[width=1\linewidth]{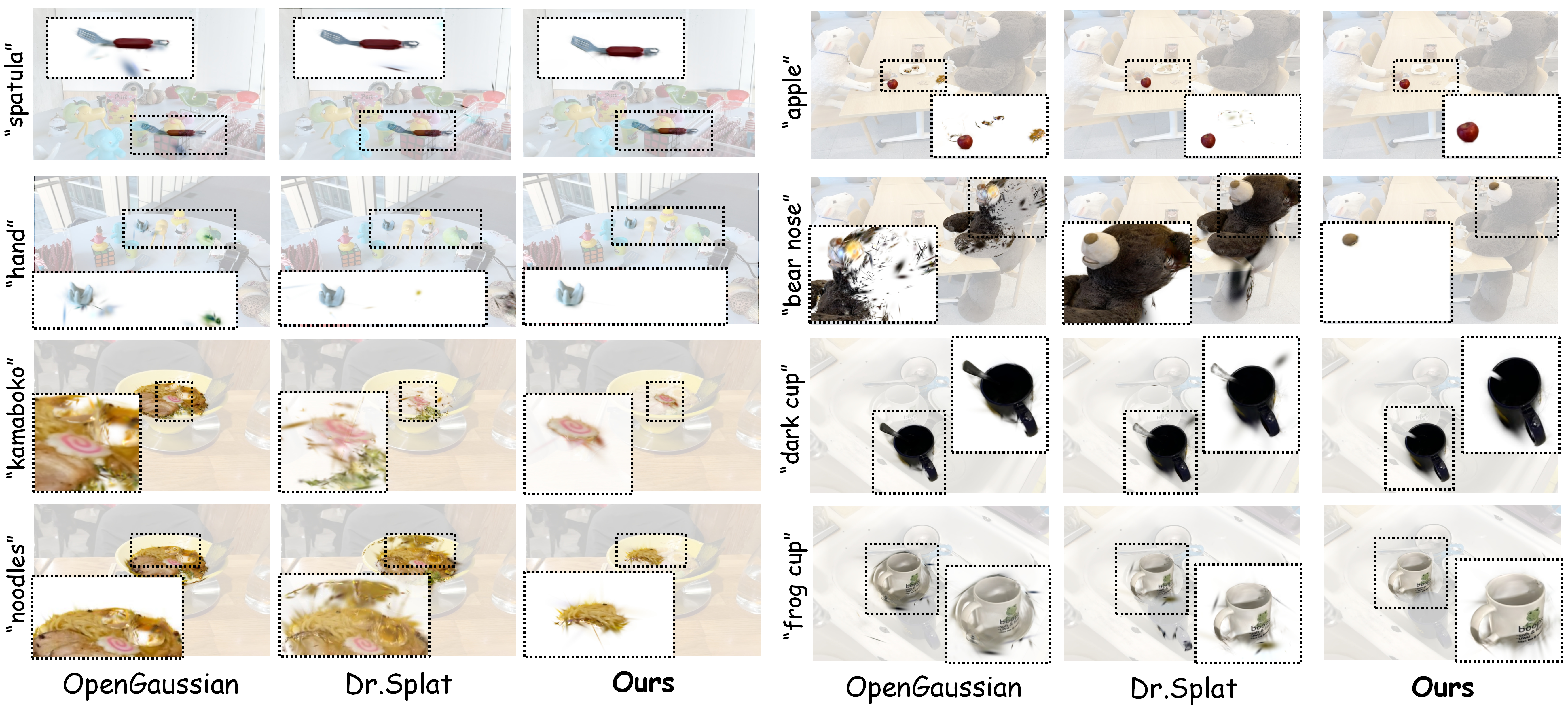}
	\caption{Qualitative results on object selection from the LERF dataset. OpenGaussian fails to separate nearby objects or maintain sharp boundaries, while Dr.Splat struggles to capture fine-grained details. In contrast, our method correctly interprets fine-grained instructions to generate precise selections with well-defined boundaries.}
	\label{fig:my_figure03}
\end{figure*}
\begin{table*}[t!]
	\centering
	\small
	\caption{This caption compares computational resources for the LERF \texttt{figurines} scene, including per-scene optimization time, peak VRAM use, and storage for CLIP features. By decoupling semantics from geometry as the only extrinsic method, our approach achieves high efficiency, cutting CLIP feature storage from gigabytes to megabytes and using the least amount of VRAM. Note that \enquote{--} marks methods that do not use CLIP features.}
	\label{tab2}
	\renewcommand{\arraystretch}{1.0} 
	\setlength{\tabcolsep}{8pt} 
	\begin{tabular}{l c c c c c c c} 
		\toprule
		Method & Venue & Domain & Paradigm & Scene Opt. & Train Time & CLIP F.S. & Peak VRAM \\
		\midrule
		LEGaussians & CVPR'24 & 2D & Embedding & Required & $\sim$2h & $\sim$3GB & $\sim$20 GB \\
		LangSplat & CVPR'24 & 2D & Embedding & Required & $\sim$2h & $\sim$3GB & $\sim$20 GB \\
		Feature-3DGS & CVPR'24 & 2D & Embedding & Required & $\sim$1h & $\sim$3GB & $\sim$26 GB \\
		GS-Grouping & ECCV'24 & 2D & Embedding & Required & $\sim$1h & -- & $\sim$28 GB \\
		GOI & MM'24 & 2D & Embedding & Required & $\sim$1h & -- & $\sim$24 GB \\
		3DVLGS & ICML'25 & 2D & Embedding & Required & $\sim$1h & $\sim$3GB & $\sim$24 GB \\
		Occam's LGS & BMVC'25 & 2D & Embedding & \textbf{None} & \textbf{None} & $\sim$3GB & \textbf{$\sim$12 GB} \\
		
		OpenGaussian & NIPS'25 & \textbf{3D} & Embedding & Required & $\sim$1h & $\sim$3GB & $\sim$22 GB \\
		InstanceGaussian & CVPR'25 & \textbf{3D} & Embedding & Required & $\sim$2h & $\sim$3GB & $\sim$24 GB \\
		Segment-then-Splat & NIPS'25 & \textbf{3D} & Embedding & Required & $\sim$1h & $\sim$3GB & $\sim$24 GB \\
		LaGa & ICML'25 & \textbf{3D} & Embedding & Required & $\sim$2h & $\sim$3GB & $\sim$24 GB \\
		Dr.Splat & CVPR'25 & \textbf{3D} & Embedding & \textbf{None} & $\sim$1h & $\sim$3GB & $\sim$24 GB \\
		LUDVIG & ICCV'25 & \textbf{3D} & Embedding & \textbf{None} & \textbf{None} & $\sim$3GB & $\sim$22 GB \\
		Ours & -- & \textbf{3D} & \textbf{Extrinsic} & \textbf{None} & \textbf{None} & \textbf{$\sim$3MB} & \textbf{$\sim$8 GB} \\
		\bottomrule
	\end{tabular}
\end{table*}

\section{Experiments}
\subsection{Open-Vocabulary Object Selection in 3D Space}
\textbf{Settings} 
\textbf{1)~Task.} Given a text query as input, the task is to produce multi-view renderings of the semantically corresponding 3D instance(s). First, the textual feature of the input query is extracted using the CLIP model. Then, cosine similarity is computed between the query feature and the textual features of each instance, and the most similar instance(s) are selected. Finally, all 3D Gaussian points belonging to the selected instances are rendered into multi-view images through the 3DGS rasterization pipeline. 
\textbf{2)~Baselines.} We compare our method with several recent representative approaches~\citep{jun-seong_dr_2025,wu2024opengaussian,qin_langsplat_2024,shi_language_2024,li_instancegaussian_2025,zhou_feature_2024,ye_gaussian_2024,qu_goi_2024,marrie2025ludvig,cen_tackling_2025,Cheng_2025_BMVC}. These approaches fall into the two primary categories of point-based and pixel-based methods. To provide a clear comparison, we detail the comparative aspects such as training time and search thresholds for these methods in Tab.~\ref{tab2}.
\textbf{3)~Dataset.} We adopt the LERF~\citep{lerf2023} dataset, annotated by LangSplat. This dataset consists of multi-view images capturing 3D scenes and provides ground-truth 2D annotations for texture-level queries. For a fair comparison, we use the same predefined query texts as in OpenGaussian.
\begin{figure*}[!t]
	\centering
	\includegraphics[width=1\linewidth]{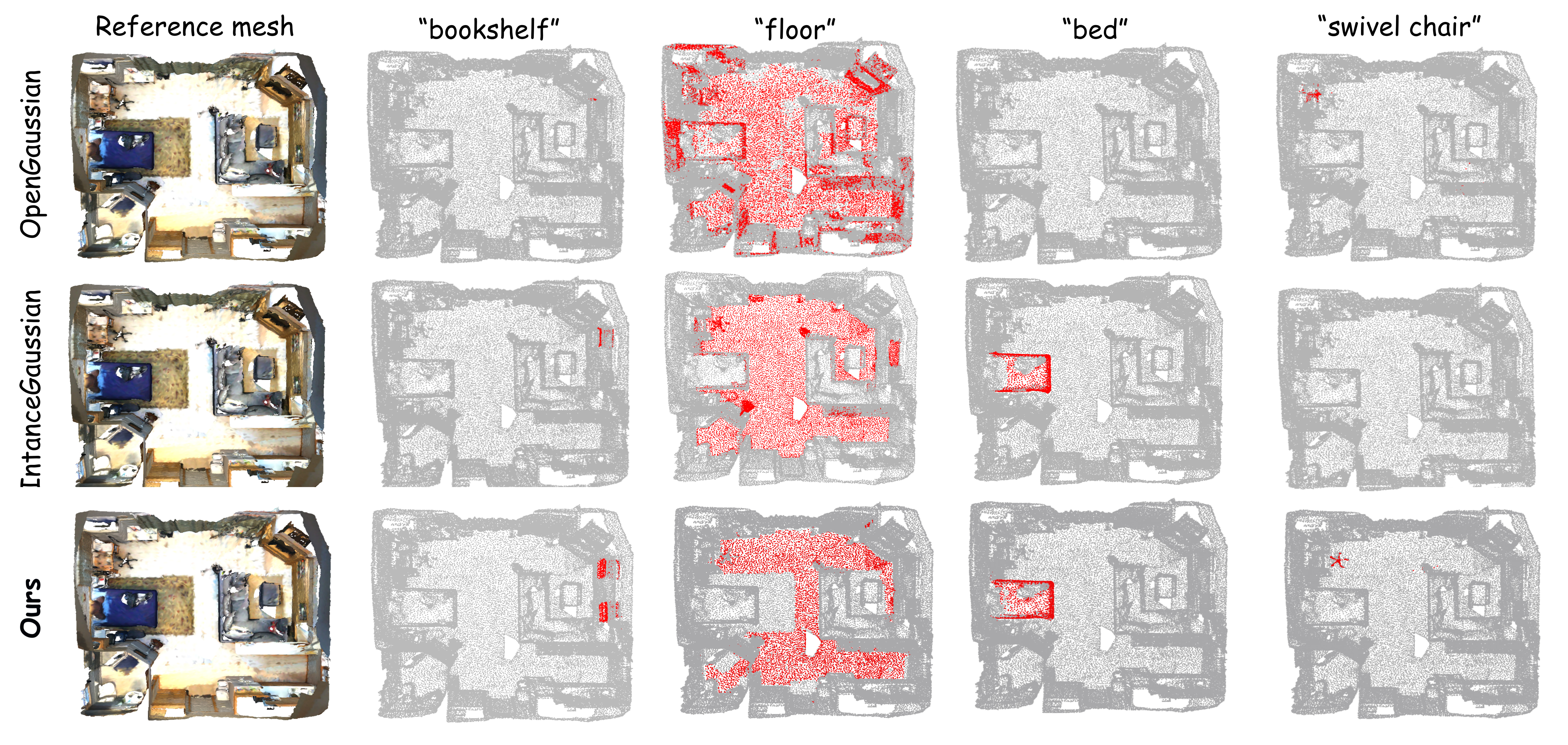}
	\caption{Qualitative results of our 3D object segmentation on the ScanNet dataset. OpenGaussian and InstanceGaussian rely on matching CLIP features extracted from 2D images. This approach is susceptible to feature inconsistencies arising from different mask viewpoints, often leading to incorrect matches (e.g., for the bed and chair). In contrast, our method achieves accurate 3D segmentation with sharp and well-defined boundaries.}
	\label{fig:my_figure04}
\end{figure*}
\begin{table}[t!]
	\centering
	\footnotesize 
	\caption{mIoU results for open-vocabulary object selection in 3D space on the LERF dataset. \textbf{Bold}/\underline{Underline} indicates the best/second-best performance per category.}
	\label{tab1}
	\renewcommand{\arraystretch}{1.0}
	\setlength{\tabcolsep}{5pt} 
	\begin{tabular}{l c c c c c} 
		\toprule
		Method & Ramen & Teatime & Figurines & Waldo & Mean \\
		\midrule
		
		\multicolumn{6}{l}{\textit{\textbf{2D Methods}}} \\
		LEGaussians & 46.0 & 60.3 & 40.8 & 39.4 & 46.6 \\
		LangSplat & 51.2 & 65.1 & 44.7 & 44.5 & 51.4 \\
		Feature-3DGS & 43.7 & 58.8 & 40.5 & 39.6 & 45.7 \\
		GS-Grouping & 45.5 & 60.9 & 40.0 & 38.7 & 46.3 \\
		GOI & \underline{52.6} & 63.7 & 44.5 & 41.4 & 50.6 \\
		Occam's LGS & 51.0 & \underline{70.2} & \textbf{58.6} & \textbf{65.3} & \underline{61.3} \\
		3DVLGS & \textbf{61.4} & \textbf{73.5} & \underline{58.1} & \underline{54.8} & \textbf{62.0} \\
		
		\midrule
		\multicolumn{6}{l}{\textit{\textbf{3D Training-based}}} \\
		LangSplat-m & 6.1 & 16.6 & 8.3 & 8.3 & 9.8 \\
		LEGaussians-m & 15.8 & 19.3 & 18.0 & 11.8 & 16.2 \\
		OpenGaussian & 31.0 & 60.4 & 39.3 & 22.7 & 38.4 \\
		InstanceGaussian & 24.6 & 63.4 & 45.5 & 29.2 & 40.7 \\
		Dr.Splat(Top-40) & 24.7 & 57.2 & \underline{53.4} & 39.1 & 43.6 \\
		Segment-then-Splat & \underline{54.8} &  \underline{63.5} & 49.8 & \underline{40.7} & \underline{52.0} \\ 
		LAGA & \textbf{55.6} & \textbf{70.9} & \textbf{64.1} & \textbf{65.6} & \textbf{64.0} \\
		
		\midrule
		\multicolumn{6}{l}{\textit{\textbf{3D Training-free}}} \\
		LUDVIG & \underline{42.3} & \underline{58.6} & \underline{58.0} & \textbf{42.8} & \underline{50.4} \\
		\textbf{Ours} & \textbf{45.6} & \textbf{64.4} & \textbf{66.4} & \underline{40.9} & \textbf{54.3} \\ 
		\bottomrule
	\end{tabular}
\end{table}
\noindent\textbf{Results} 
\textbf{1) Quantitative Evaluation.} As shown in Tab.~\ref{tab2}, our extrinsic, decoupled architecture eliminates per-scene optimization, leading to a $\sim$1000x reduction in feature storage and a significantly lower VRAM footprint. As shown in Tab.~\ref{tab1}, our method achieves a new state-of-the-art (SOTA) result, outperforming the previous best-performing method by 3.9 mIoU. This highlights the efficacy of our decoupled architecture in maximizing both segmentation accuracy and resource efficiency.
\textbf{2)~Qualitative Evaluation.} 
As illustrated in Fig.~\ref{fig:my_figure03}, competing methods expose the inherent flaws of current feature-embedding paradigms. OpenGaussian aggregates semantically relevant Gaussians into singular representations, conflating part-level semantics and neglecting neutral points, which yields chaotic boundaries and irrelevant selections (e.g., spatula, apple). While Dr.Splat compresses CLIP features to reduce bloat, this degrades linguistic precision and retains the flaw of fusing multiple semantic identities, failing on fine-grained queries (e.g., \enquote{bear nose}, \enquote{noodles}). In contrast, our extrinsic architecture inherently avoids feature entanglement. By operating on semantic objects, utilizing uncompressed CLIP representations for maximal linguistic fidelity, and explicitly handling neutral points, our approach guarantees precise semantics and sharp geometric boundaries.
\begin{table}[t!]
	\centering
	\footnotesize 
	\caption{Quantitative results for open-vocabulary 3D semantic segmentation on the ScanNet dataset. \textbf{Bold}/\underline{Underline} indicates the best/second-best performance per category.}
	\label{tab:scannet_results_revised}
	\renewcommand{\arraystretch}{1.0} 
	
	\setlength{\tabcolsep}{3pt} 
	\begin{tabular}{l cccccc} 
		\toprule
		\multirow{2}{*}{Method} & \multicolumn{2}{c}{19 classes} & \multicolumn{2}{c}{15 classes} & \multicolumn{2}{c}{10 classes} \\
		& mIoU$\uparrow$ & mAcc$\uparrow$ & mIoU$\uparrow$ & mAcc$\uparrow$ & mIoU$\uparrow$ & mAcc$\uparrow$ \\
		\midrule
		LangSplat-m & 3.8 & 9.1 & 5.4 & 13.2 & 8.4 & 22.1 \\
		LEGaussians-m & 1.6 & 7.9 & 4.6 & 16.1 & 7.7 & 24.9 \\
		OpenGaussian & 24.7 & 41.5 & 30.1 & 48.3 & 38.3 & 55.2 \\
		InstanceGaussian & \underline{40.7} & \underline{54.0} & \underline{42.5} & 59.1 & 47.9 & 64.0 \\
		Dr.Splat(Top-40) & 29.6 & 47.7 & 38.2 & \underline{60.4} & \underline{50.8} & \underline{73.5} \\
		LAGA & 32.5 & 49.1 & 35.5 & 53.5 &  42.6 & 63.2 \\
		LUDVIG & 33.9 & 51.4 & 37.4 & 57.2 & 46.4 & 66.2 \\
		\textbf{Ours} & \textbf{45.5} & \textbf{58.4} & \textbf{47.2} & \textbf{61.7} & \textbf{53.7} & \textbf{74.9} \\
		\bottomrule
	\end{tabular}
\end{table}

\subsection{Open-Vocabulary 3D Semantic Segmentation}
\textbf{Settings} \textbf{1) Task.} The objective is to automatically extract 3D Gaussian points corresponding to input class names (e.g., wall, chair, table). The segmented Gaussian points are then converted into a point cloud to be evaluated against the ground-truth annotated point cloud. To ensure a precise correspondence between the converted point cloud and the ground truth, we disable the 3D Gaussian densification process during training.  
\textbf{2) Baselines.} Consistent with the object selection task, we compare our method against several recently proposed approaches~\citep{marrie2025ludvig,jun-seong_dr_2025,li_instancegaussian_2025,wu2024opengaussian,cen_tackling_2025}. LangSplat-m and LEGaussians-m are adaptations of existing pixel-based methods~\citep{shi_language_2024,qin_langsplat_2024}, specifically modified to perform direct 3D referring operations. As this task requires a direct understanding of 3D points, pixel-based methods are not applicable and are therefore excluded from our comparison. 
\textbf{3) Dataset.} We employ the ScanNet~\citep{dai2017scannet} dataset, a  benchmark comprising indoor scene data with calibrated RGB-D images and 3D point clouds annotated with ground-truth semantic labels. For a fair comparison, we adopt the same scenes and evaluation categories used in OpenGaussian.

\noindent \textbf{Results} \textbf{1) Quantitative Analysis.} Tab.~\ref{tab:scannet_results_revised} shows the performance on the ScanNet dataset using text queries for 19, 15, and 10 of its classes. The results show that our method consistently achieves SOTA segmentation performance across all scenes relative to the baselines. This consistent leadership strongly suggests a more robust semantic understanding, validating our hypothesis that our extrinsic, VLM-driven approach provides a more stable representation than competing embedding methods struggling with 3D semantic inconsistencies.
\textbf{2) Qualitative Analysis.} Qualitative results are presented in Fig.~\ref{fig:my_figure04}. In complex scenes from ScanNet, both OpenGaussian and InstanceGaussian frequently exhibit incorrect matches, which limits their accuracy. This limitation arises from their reliance on matching masked CLIP image features, as semantic inconsistencies across different viewing angles make it difficult for such methods to achieve high-precision results. In contrast, our method instead leverages a VLM to distill an object's varied and often-corrupted visual appearances into a set of canonical textual hypotheses, achieving a robust semantic understanding that is invariant to occlusion.
\begin{table}[t!] 
	\centering
	\small 
	\renewcommand{\arraystretch}{1.1} 
	\setlength{\tabcolsep}{10pt} 
	\caption{Ablation on neutral point processing. We evaluate the impact of our two-stage filtering on the LERF dataset.}
	\label{tab:ablation_neutral}
	\begin{tabular}{c l c} 
		\toprule
		Case & Method & mIoU$\uparrow$ \\
		\midrule
		\#1 & Initial Grouping & 53.0 \\
		\#2 & + Opacity Filter & 52.6 \\
		\#3 & + Entropy Filter & 53.2 \\
		\#4 & + Entropy \& Opacity Filters (Ours) & \textbf{54.3} \\
		\bottomrule
	\end{tabular}
\end{table}

\subsection{Ablation Study}
\textbf{Neutral Point Processing.} We ablate our neutral point processing on the LERF dataset with results in Tab.~\ref{tab:ablation_neutral}. Case \#1 is the baseline without any filtering. Case \#2 applies only the opacity filter, and Case \#3 applies only the entropy filter. Case \#4 introduces our full model, which incorporates both filters. As shown, entropy filtering alone provides a minor gain by suppressing noise, but the opacity filter alone can be aggressive and inadvertently remove valid foreground points. Combining both filters resolves this issue, achieving the highest mIoU. This demonstrates that both stages are essential for the final performance.

\noindent \textbf{Instance Feature Extraction.}
To demonstrate the advantages of using a VLM for language feature extraction, we compare our approach with three baselines derived from the CLIP image encoder. 
Case \#1 uses the feature from the single view with the largest mask area.
Case \#2 averages features from all valid views. 
Case \#3 first renders the class foreground points onto each view and computes the IoU between the rendered foreground masks and candidate masks. We then discard the low-IoU masks and average the features of the remaining ones. 
The results are presented in Tab.~\ref{tab:ablation_feature}.
Single-view methods struggle to capture comprehensive semantics, while multi-view averaging methods often yield ambiguous features due to occlusions. Although filtering-based methods significantly improve matching accuracy, they require a rendering pass for each view, which incurs high runtime costs. Moreover, these methods can obscure discriminative details due to feature discrepancies across different views (see Appendix for details). In contrast, our VLM-based method distills these multi-view cues into a consistent textual representation, effectively capturing the nuanced attributes required for abstract queries.

\begin{table}[t!] 
	\centering
	\small 
	\renewcommand{\arraystretch}{1.1} 
	\setlength{\tabcolsep}{8pt} 
	\caption{Ablation on feature extraction. We compare VLM-based text distillation against CLIP image baselines.}
	\label{tab:ablation_feature}
	\begin{tabular}{c c c c} 
		\toprule
		Case & Feature Source & View Aggregation & mIoU$\uparrow$ \\
		\midrule
		\#1 & Image  & Single (Max Area) & 36.9 \\
		\#2 & Image  & Average (All) & 39.2 \\
		\#3 & Image  & Average (Filtered) & 50.1 \\
		\#4 & Text & Holistic & \textbf{54.3} \\
		\bottomrule
	\end{tabular}
\end{table}


\section{Conclusion and Limitation} 
In this work, we introduced ExtrinSplat, a training-free framework that realizes the extrinsic paradigm for open-vocabulary 3D Gaussian understanding. 
Our architecture realizes this paradigm by constructing an extrinsic semantic index layer that completely separates geometry from semantics, associating purified 3D geometric groups with stable textual hypotheses to enable efficient open-vocabulary matching.
Evaluations on multiple benchmarks confirm that ExtrinSplat delivers SOTA-level performance at a fraction of the computational cost. This work demonstrates that our extrinsic paradigm is not just a viable alternative, but a more efficient and semantically robust foundation for open-vocabulary 3D Gaussian understanding. 

Despite its strong performance, our method has certain limitations: \textbf{1)}~The accuracy of our object-level grouping can be compromised by substantially inaccurate initial segmentation masks from SAM. \textbf{2)}~Rarely, the VLM may assign incorrect semantic labels to objects. Addressing these issues remains a promising direction for future work.

\section*{Acknowledgements}
This work was supported by the Natural Science Foundation of China (Grant No. 62531022), the Guangdong Provincial Key Laboratory of Ultra High Definition Immersive Media Technology (Grant No. 2024B1212010006), and the Outstanding Talents Training Fund in Shenzhen.

{
    \small
    \bibliographystyle{ieeenat_fullname}
    \bibliography{main}

@String(ICCV= {Int. Conf. Comput. Vis.})

@String(BMVC= {Brit. Mach. Vis. Conf.})

@String(ICCV  = {ICCV})

@String(BMVC  =	{BMVC})

@article{kong2025multi,
  title={Multi-modal data-efficient 3d scene understanding for autonomous driving},
  author={Kong, Lingdong and Xu, Xiang and Ren, Jiawei and Zhang, Wenwei and Pan, Liang and Chen, Kai and Ooi, Wei Tsang and Liu, Ziwei},
  journal={IEEE Transactions on Pattern Analysis and Machine Intelligence},
  year={2025},
  publisher={IEEE}
}

@article{oquab2023dinov2,
  title={Dinov2: Learning robust visual features without supervision},
  author={Oquab, Maxime and Darcet, Timoth{\'e}e and Moutakanni, Th{\'e}o and Vo, Huy and Szafraniec, Marc and Khalidov, Vasil and Fernandez, Pierre and Haziza, Daniel and Massa, Francisco and El-Nouby, Alaaeldin and others},
  journal={arXiv preprint arXiv:2304.07193},
  year={2023}
}

@article{liang2024supergseg,
  title={Supergseg: Open-vocabulary 3d segmentation with structured super-gaussians},
  author={Liang, Siyun and Wang, Sen and Li, Kunyi and Niemeyer, Michael and Gasperini, Stefano and Navab, Nassir and Tombari, Federico},
  journal={arXiv preprint arXiv:2412.10231},
  year={2024}
}

@article{sun2025cags,
  title={Cags: Open-vocabulary 3d scene understanding with context-aware gaussian splatting},
  author={Sun, Wei and Zhou, Yanzhao and Jiao, Jianbin and Li, Yuan},
  journal={arXiv preprint arXiv:2504.11893},
  year={2025}
}

@article{yin2025semantic,
  title={Semantic Consistent Language Gaussian Splatting for Point-Level Open-vocabulary Querying},
  author={Yin, Hairong and Zhan, Huangying and Xu, Yi and Yeh, Raymond A},
  journal={arXiv preprint arXiv:2503.21767},
  year={2025}
}

@article{jiang2025votesplat,
  title={VoteSplat: Hough Voting Gaussian Splatting for 3D Scene Understanding},
  author={Jiang, Minchao and Jia, Shunyu and Gu, Jiaming and Lu, Xiaoyuan and Zhu, Guangming and Dong, Anqi and Zhang, Liang},
  journal={arXiv preprint arXiv:2506.22799},
  year={2025}
}

@inproceedings{zhu2023understanding,
  title={Understanding the robustness of 3D object detection with bird's-eye-view representations in autonomous driving},
  author={Zhu, Zijian and Zhang, Yichi and Chen, Hai and Dong, Yinpeng and Zhao, Shu and Ding, Wenbo and Zhong, Jiachen and Zheng, Shibao},
  booktitle={Proceedings of the IEEE/CVF Conference on Computer Vision and Pattern Recognition},
  pages={21600--21610},
  year={2023}
}

@inproceedings{chen2024sugar,
  title={Sugar: Pre-training 3d visual representations for robotics},
  author={Chen, Shizhe and Garcia, Ricardo and Laptev, Ivan and Schmid, Cordelia},
  booktitle={Proceedings of the IEEE/CVF Conference on Computer Vision and Pattern Recognition},
  pages={18049--18060},
  year={2024}
}

@inproceedings{song2025robospatial,
  title={Robospatial: Teaching spatial understanding to 2d and 3d vision-language models for robotics},
  author={Song, Chan Hee and Blukis, Valts and Tremblay, Jonathan and Tyree, Stephen and Su, Yu and Birchfield, Stan},
  booktitle={Proceedings of the Computer Vision and Pattern Recognition Conference},
  pages={15768--15780},
  year={2025}
}

@inproceedings{Cheng_2025_BMVC,
author    = {Jiahuan Cheng and Jan-Nico Zaech and Luc Van Gool and Danda Pani Paudel},
title     = {Occam’s LGS: An Efficient Approach for Language Gaussian Splatting},
booktitle = {36th British Machine Vision Conference 2025, {BMVC} 2025, Sheffield, UK, November 24-27, 2025},
publisher = {BMVA},
year      = {2025}
}

@inproceedings{marrie2025ludvig,
    title={LUDVIG: Learning-Free Uplifting of 2D Visual Features to Gaussian Splatting Scenes},
    author={Marrie, Juliette and Menegaux, Romain and Arbel, Michael and Larlus, Diane and Mairal, Julien},
    booktitle={Proceedings of the IEEE/CVF International Conference on Computer Vision (ICCV)},
    year={2025}
}

@misc{comanici2025gemini25pushingfrontier,
	title={{Gemini 2.5: Pushing the Frontier with Advanced Reasoning, Multimodality, Long Context, and Next Generation Agentic Capabilities}}, 
	author={Comanici, Gheorghe and others},
	year={2025},
	eprint={2507.06261},
	howpublished = {arXiv preprint arXiv:2507.06261},
	archivePrefix={arXiv},
	primaryClass={cs.CL},
}

@misc{chen2024internvlscalingvisionfoundation,
	title={InternVL: Scaling up Vision Foundation Models and Aligning for Generic Visual-Linguistic Tasks}, 
	author={Zhe Chen and Jiannan Wu and Wenhai Wang and Weijie Su and Guo Chen and Sen Xing and Muyan Zhong and Qinglong Zhang and Xizhou Zhu and Lewei Lu and Bin Li and Ping Luo and Tong Lu and Yu Qiao and Jifeng Dai},
	year={2024},
	howpublished = {arXiv preprint arXiv:2312.14238},
	eprint={2312.14238},
	archivePrefix={arXiv},
	primaryClass={cs.CV},
}

@misc{kerbl_3d_2023,
	title     = {{3D} {Gaussian} {Splatting} for {Real}-{Time} {Radiance} {Field} {Rendering}},
	publisher = {arXiv},
	author    = {Kerbl, Bernhard and Kopanas, Georgios and Leimkühler, Thomas and Drettakis, George},
	howpublished = {arXiv preprint arXiv:2308.04079},
	year      = {2023},
	note      = {arXiv:2308.04079},
}

@inproceedings{zhou_feature_2024,
	title     = {Feature {3Dgs}: {Supercharging} 3d gaussian splatting to enable distilled feature fields},
	shorttitle= {Feature 3dgs},
	booktitle = {Proceedings of the {IEEE}/{CVF} {Conference} on {Computer} {Vision} and {Pattern} {Recognition}},
	author    = {Zhou, Shijie and Chang, Haoran and Jiang, Sicheng and Fan, Zhiwen and Zhu, Zehao and Xu, Dejia and Chari, Pradyumna and You, Suya and Wang, Zhangyang and Kadambi, Achuta},
	year      = {2024},
	pages     = {21676--21685},
}

@article{wu2024opengaussian,
	title   = {Opengaussian: Towards point-level 3d gaussian-based open vocabulary understanding},
	author  = {Wu, Yanmin and Meng, Jiarui and Li, Haijie and Wu, Chenming and Shi, Yahao and Cheng, Xinhua and Zhao, Chen and Feng, Haocheng and Ding, Errui and Wang, Jingdong and others},
	journal = {Advances in Neural Information Processing Systems},
	volume  = {37},
	pages   = {19114--19138},
	year    = {2024}
}

@inproceedings{shen2024flashsplat,
	title        = {Flashsplat: 2d to 3d gaussian splatting segmentation solved optimally},
	author       = {Shen, Qiuhong and Yang, Xingyi and Wang, Xinchao},
	booktitle    = {European Conference on Computer Vision},
	pages        = {456--472},
	year         = {2024},
	organization = {Springer}
}

@inproceedings{li_instancegaussian_2025,
	title     = {Instancegaussian: {Appearance}-semantic joint gaussian representation for {3D} instance-level perception},
	shorttitle= {Instancegaussian},
	booktitle = {Proceedings of the {Computer} {Vision} and {Pattern} {Recognition} {Conference}},
	author    = {Li, Haijie and Wu, Yanmin and Meng, Jiarui and Gao, Qiankun and Zhang, Zhiyao and Wang, Ronggang and Zhang, Jian},
	year      = {2025},
	pages     = {14078--14088},
}

@inproceedings{shi_language_2024,
	title     = {Language embedded {3D} gaussians for open-vocabulary scene understanding},
	booktitle = {Proceedings of the {IEEE}/{CVF} {Conference} on {Computer} {Vision} and {Pattern} {Recognition}},
	author    = {Shi, Jin-Chuan and Wang, Miao and Duan, Hao-Bin and Guan, Shao-Hua},
	year      = {2024},
	pages     = {5333--5343},
}

@inproceedings{qin_langsplat_2024,
	title     = {Langsplat: {3D} language gaussian splatting},
	shorttitle= {Langsplat},
	booktitle = {Proceedings of the {IEEE}/{CVF} {Conference} on {Computer} {Vision} and {Pattern} {Recognition}},
	author    = {Qin, Minghan and Li, Wanhua and Zhou, Jiawei and Wang, Haoqian and Pfister, Hanspeter},
	year      = {2024},
	pages     = {20051--20060},
}

@inproceedings{ye_gaussian_2024,
	title     = {Gaussian grouping: {Segment} and edit anything in {3D} scenes},
	shorttitle= {Gaussian grouping},
	booktitle = {European conference on computer vision},
	publisher = {Springer},
	author    = {Ye, Mingqiao and Danelljan, Martin and Yu, Fisher and Ke, Lei},
	year      = {2024},
	pages     = {162--179},
}

@inproceedings{qu_goi_2024,
	address   = {Melbourne VIC Australia},
	title     = {{GOI}: {Find} {3D} gaussians of interest with an optimizable open-vocabulary semantic-space hyperplane},
	isbn      = {979-8-4007-0686-8},
	shorttitle= {{GOI}},
	doi       = {10.1145/3664647.3680852},
	language  = {en},
	booktitle = {Proceedings of the 32nd {ACM} {International} {Conference} on {Multimedia}},
	publisher = {ACM},
	author    = {Qu, Yansong and Dai, Shaohui and Li, Xinyang and Lin, Jianghang and Cao, Liujuan and Zhang, Shengchuan and Ji, Rongrong},
	year      = {2024},
}

@inproceedings{jun-seong_dr_2025,
	title     = {Dr. splat: {Directly} referring {3D} gaussian splatting via direct language embedding registration},
	shorttitle= {Dr. splat},
	booktitle = {Proceedings of the {Computer} {Vision} and {Pattern} {Recognition} {Conference}},
	author    = {Jun-Seong, Kim and Kim, GeonU and Yu-Ji, Kim and Wang, Yu-Chiang Frank and Choe, Jaesung and Oh, Tae-Hyun},
	year      = {2025},
	pages     = {14137--14146},
}

@misc{cen_tackling_2025,
	title     = {Tackling view-dependent semantics in {3D} language gaussian splatting},
	doi       = {10.48550/arXiv.2505.24746},
	publisher = {arXiv},
	howpublished = {arXiv preprint arXiv:2505.24746},
	author    = {Cen, Jiazhong and Zhou, Xudong and Fang, Jiemin and Wen, Changsong and Xie, Lingxi and Zhang, Xiaopeng and Shen, Wei and Tian, Qi},
	year      = {2025},
}

@inproceedings{radford_learning_2021,
	title     = {Learning transferable visual models from natural language supervision},
	booktitle = {International conference on machine learning},
	publisher = {PmLR},
	author    = {Radford, Alec and Kim, Jong Wook and Hallacy, Chris and Ramesh, Aditya and Goh, Gabriel and Agarwal, Sandhini and Sastry, Girish and Askell, Amanda and Mishkin, Pamela and Clark, Jack},
	year      = {2021},
	pages     = {8748--8763},
}

@misc{zhang_dino_2022,
	title     = {{DINO}: {DETR} with improved {DeNoising} anchor boxes for end-to-end object detection},
	shorttitle= {{DINO}},
	doi       = {10.48550/arXiv.2203.03605},
howpublished = {arXiv preprint arXiv:2203.03605},
publisher = {arXiv},
author    = {Zhang, Hao and Li, Feng and Liu, Shilong and Zhang, Lei and Su, Hang and Zhu, Jun and Ni, Lionel M. and Shum, Heung-Yeung},
year      = {2022},
}

@inproceedings{kirillov_segment_2023,
	title     = {Segment anything},
	booktitle = {Proceedings of the {IEEE}/{CVF} international conference on computer vision},
	author    = {Kirillov, Alexander and Mintun, Eric and Ravi, Nikhila and Mao, Hanzi and Rolland, Chloe and Gustafson, Laura and Xiao, Tete and Whitehead, Spencer and Berg, Alexander C. and Lo, Wan-Yen},
	year      = {2023},
	pages     = {4015--4026},
}

@inproceedings{munkberg2022extracting,
	title={Extracting triangular 3d models, materials, and lighting from images},
	author={Munkberg, Jacob and Hasselgren, Jon and Shen, Tianchang and Gao, Jun and Chen, Wenzheng and Evans, Alex and M{\"u}ller, Thomas and Fidler, Sanja},
	booktitle={Proceedings of the IEEE/CVF Conference on Computer Vision and Pattern Recognition},
	pages={8280--8290},
	year={2022}
}

@inproceedings{lu_scaffold-gs_2024,
	title     = {Scaffold-gs: {Structured} {3D} gaussians for view-adaptive rendering},
	shorttitle= {Scaffold-gs},
	booktitle = {Proceedings of the {IEEE}/{CVF} {Conference} on {Computer} {Vision} and {Pattern} {Recognition}},
	author    = {Lu, Tao and Yu, Mulin and Xu, Linning and Xiangli, Yuanbo and Wang, Limin and Lin, Dahua and Dai, Bo},
	year      = {2024},
	pages     = {20654--20664},
}

@inproceedings{dai2017scannet,
	title     = {Scannet: Richly-annotated 3d reconstructions of indoor scenes},
	author    = {Dai, Angela and Chang, Angel X and Savva, Manolis and Halber, Maciej and Funkhouser, Thomas and Nie{\ss}ner, Matthias},
	booktitle = {Proceedings of the IEEE conference on computer vision and pattern recognition},
	pages     = {5828--5839},
	year      = {2017}
}

@inproceedings{lerf2023,
	author = {Kerr, Justin* and Kim, Chung Min* and Goldberg, Ken and Kanazawa, Angjoo and Tancik, Matthew},
	title = {LERF: Language Embedded Radiance Fields},
	booktitle = {International Conference on Computer Vision (ICCV)},
	year = {2023},
}

@article{fang2023robust,
	title={Robust grasping across diverse sensor qualities: The GraspNet-1Billion dataset},
	author={Fang, Hao-Shu and Gou, Minghao and Wang, Chenxi and Lu, Cewu},
	journal={The International Journal of Robotics Research},
	year={2023},
	publisher={SAGE Publications Sage UK: London, England}
}

@misc{lu_segment_2025,
	title     = {Segment then splat: {A} unified approach for {3D} open-vocabulary segmentation based on gaussian splatting},
	shorttitle= {Segment then {Splat}},
	doi       = {10.48550/arXiv.2503.22204},
	publisher = {arXiv},
	howpublished = {arXiv preprint arXiv:2503.22204},
	author    = {Lu, Yiren and Zhou, Yunlai and Qiao, Yiran and Song, Chaoda and Liang, Tuo and Ma, Jing and Yin, Yu},
	year      = {2025}
}

@inproceedings{videnovic2025distractor,
	title={A distractor-aware memory for visual object tracking with sam2},
	author={Videnovic, Jovana and Lukezic, Alan and Kristan, Matej},
	booktitle={Proceedings of the Computer Vision and Pattern Recognition Conference},
	pages={24255--24264},
	year={2025}
}

@misc{li_gradiseg_2024,
	title     = {{GradiSeg}: {Gradient}-guided gaussian segmentation with enhanced {3D} boundary precision},
	shorttitle= {{GradiSeg}},
	doi       = {10.48550/arXiv.2412.00392},
	publisher = {arXiv},
	author    = {Li, Zehao and Han, Wenwei and Cai, Yujun and Jiang, Hao and Bi, Baolong and Gao, Shuqin and Zhao, Honglong and Wang, Zhaoqi},
	howpublished = {arXiv preprint arXiv:2412.00392},
	year      = {2024}
}

@inproceedings{zhang2025cobgs,
	title     = {COB-GS: Clear Object Boundaries in 3DGS Segmentation Based on Boundary-Adaptive Gaussian Splitting},
	author    = {Zhang, Jiaxin and Jiang, Junjun and Chen, Youyu and Jiang, Kui and Liu, Xianming},
	booktitle ={Proceedings of the Computer Vision and Pattern Recognition Conference},
	year      = {2025}
}
}

\clearpage
\setcounter{page}{1}
\maketitlesupplementary
\section{Implementation Details}
\subsection{Model Implementation Details}
\label{mid}
\textbf{Data Preparation.} Initially, we employ SAM with grid-based point prompting to acquire initial static object masks at varying granularities from the first input frame, $I_0$. Subsequently, these masks extracted from $I_0$ are utilized by the DAM2SAM~\citep{videnovic2025distractor} model to track the corresponding objects throughout the entire image sequence.

To ensure all objects appearing throughout the sequence are captured, we introduce a periodic new-object detection mechanism. This check is performed at a fixed interval of $\Delta t = 10$ frames. At each check, we first compute the total area of all tracked masks in the current frame, $A_t$. We then trigger a full re-segmentation on this frame using SAM to get a candidate mask area, $A_{cand}$. A potential new object event is flagged if the ratio $A_t / A_{cand}$ falls below a threshold $\tau_{area} = 0.9$. When triggered, we identify a mask from the candidate set as a \enquote{new} object if its maximum Intersection over Union (IoU) with any existing tracked mask is below a threshold of $\tau_{iou} = 0.6$. Once identified, these new objects are added to the tracking pool and propagated by DAM2SAM henceforth.

Existing research~\citep{lu_segment_2025} suggests that such a detection mechanism can introduce two potential drawbacks: (1) tracking failures for some objects, resulting in incomplete object tracks, and (2) re-appearing objects being misidentified as new after their tracking has been lost, leading to a single object being assigned multiple instance IDs. Our model, however, does not need to overcome these issues during the data preparation stage.

Regarding the first issue,  we simply discard views with empty masks (i.e., where object tracking has failed) during our object-level grouping stage. As demonstrated in Appendix \ref{aas}, our model achieves robust performance even with a reduced number of views per object. Consequently, this issue has a negligible impact on the overall model accuracy.

Regarding the second issue, the emergence of multiple instances for a single object is handled by our matching process. The matching between open-vocabulary queries and instance point clusters is a one-to-many operation based on similarity. In the event of multiple matches, we take the union of their results as the final output. Therefore, the presence of multiple instances for the same object does not degrade the final matching accuracy.

In summary, our model imposes minimal requirements on the data preparation stage and functions effectively even with partial mask information for each object. This demonstrates the robustness of our approach to imperfections in the input data.

\noindent \textbf{Object-Level Grouping.} The object-level grouping process is accomplished within a single forward rendering pass. In our implementation, we simply accumulate the contribution weights of all participating 3D Gaussians during the forward pass of the 3D Gaussian Splatting render. Throughout this process, the contribution weight of each Gaussian is naturally aggregated, obviating the need for auxiliary data structures or redundant computations. By leveraging the highly optimized volumetric projection inherent to 3D Gaussian Splatting, our method achieves exceptional computational efficiency while maintaining semantic coherence.
For the subsequent neutral point processing, we use fixed thresholds across all experiments to ensure robustness and consistency. The semantic entropy threshold is set to $\tau_h = 0.9$, and the opacity threshold for filtering is set to $\tau_\alpha = 0.1$. A detailed sensitivity analysis for these hyperparameters is provided in Appendix~\ref{aas}.

\noindent \textbf{Instance Feature Extraction.} 	We acquire features for each object instance as follows. First, we identify the three largest masks for the instance based on pixel area. For each selected mask, we highlight the corresponding object on the original image with a green bounding box, creating three distinct input images. These images are then processed by a VLM, which generates a set of five nouns that describe the instance.

To match an instance against a user's text query, we compute the cosine similarity between the CLIP feature embedding of the query and the CLIP embeddings of the five nouns associated with that instance. This design allows a single query to potentially match multiple instances. A match is deemed successful if the similarity score for \textit{any} of an instance's five candidate nouns surpasses a predefined threshold of $\eta = 0.9$.

The specific prompt template used to elicit these nouns from the VLM is defined as follows:

\textit{``In the images, identify the object that is enclosed by a bright green outline. Provide five distinct and appropriate nouns to describe ONLY that specific object. Return ONLY the five nouns separated by slashes (e.g., car/automobile/vehicle/motorcar/transport). Do not add any other explanatory text, titles, or formatting.''}

\begin{figure*}[t]
	\centering
	\includegraphics[width=1\linewidth]{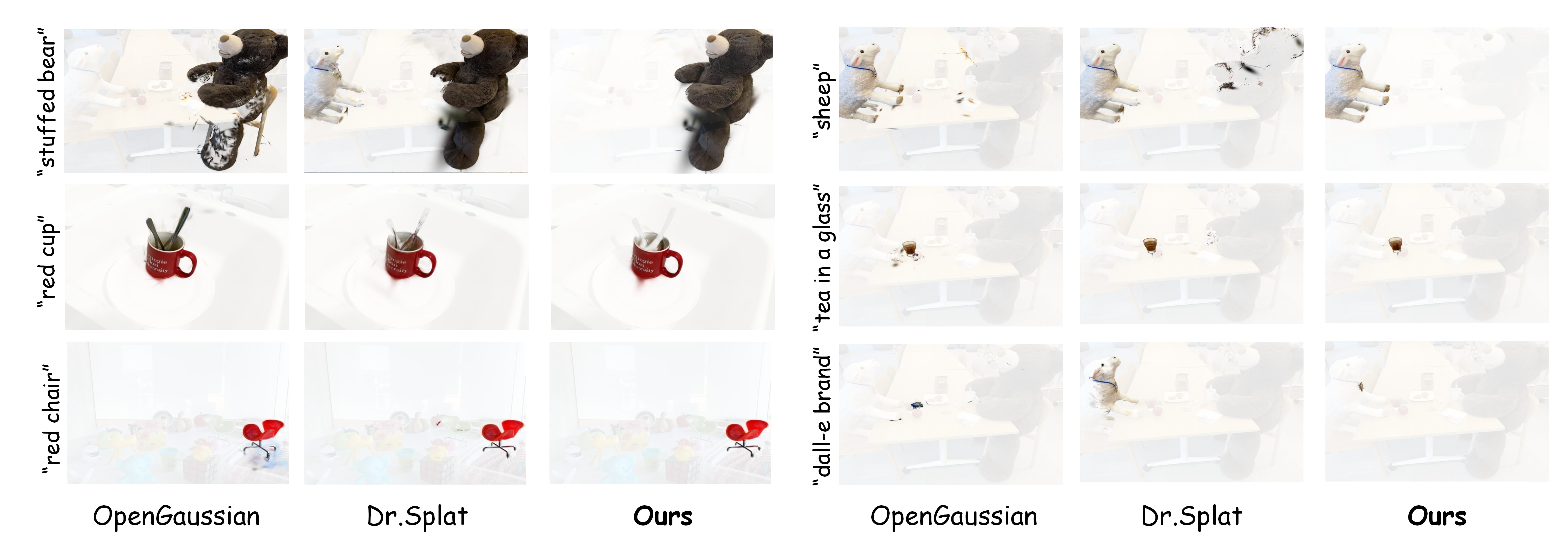}
	\caption{Additional qualitative results for open-vocabulary object selection on the LERF dataset.}
	\label{006}
\end{figure*}
\begin{figure*}[t]
	\centering
	\includegraphics[width=1\linewidth]{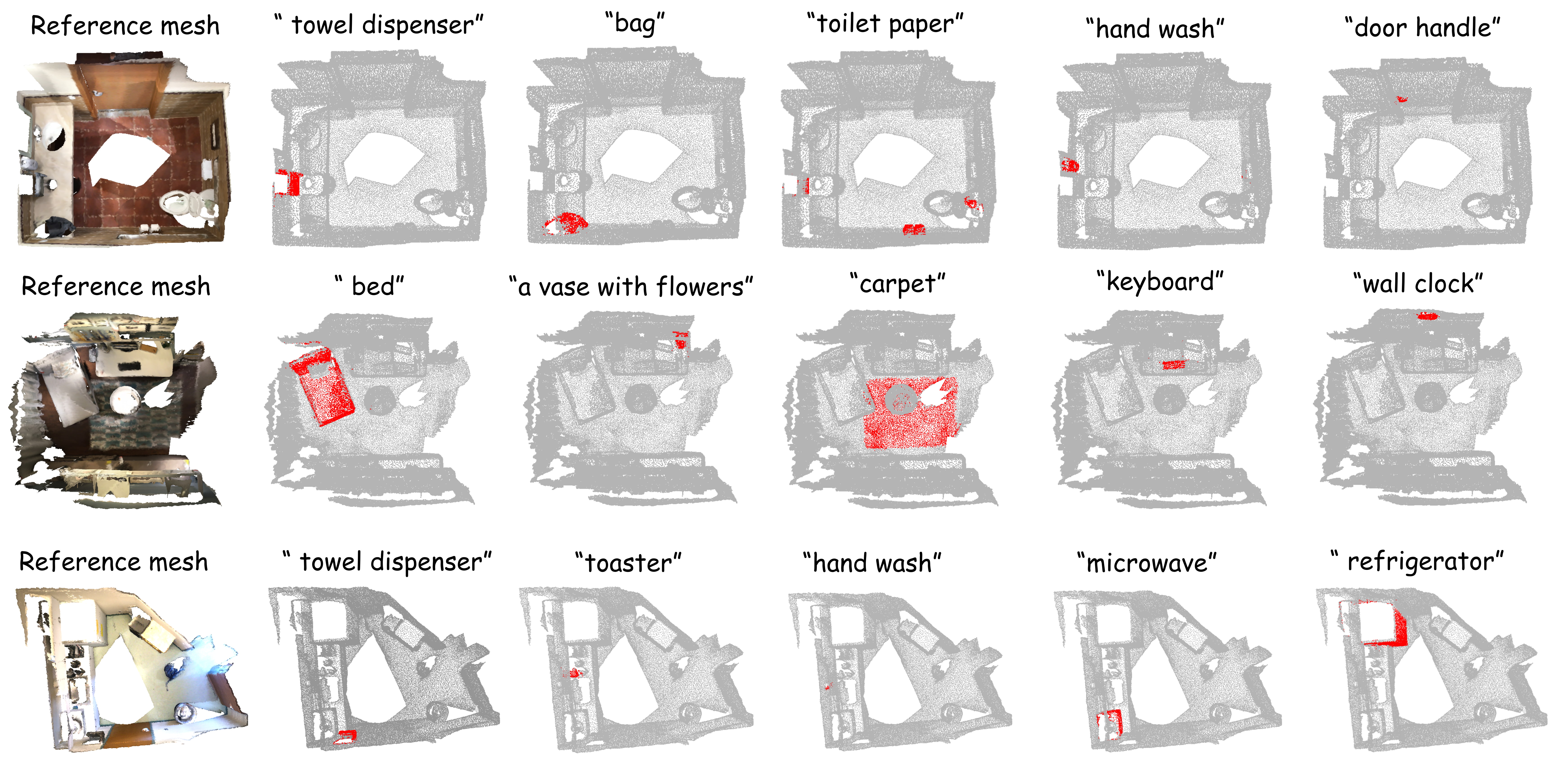}
	\caption{Additional qualitative results for open-vocabulary 3D semantic segmentation on the ScanNet dataset.}
	\label{007}
\end{figure*}

\subsection{Evaluation Details}
\noindent \textbf{LERF Dataset Evaluation}
We evaluate our model on the LERF dataset, using annotations from LangSplat. Due to the absence of 3D ground truth, we follow the 2D-based evaluation protocol from OpenGaussian. This protocol measures 3D understanding by computing the multi-view IoU accuracy between rendered occupancy masks from our selected 3D Gaussians and the ground-truth masks, which were manually annotated and provided by OpenGaussian for a set of text queries.

\noindent \textbf{ScanNet Dataset Evaluation}
\label{sec:sce}
For evaluation on the ScanNet dataset, we select the same 10 scenes as used in OpenGaussian: scene0000\_00, scene0062\_00, scene0070\_00, scene0097\_00, scene0140\_00, scene0200\_00, scene0347\_00, scene0400\_00, scene0590\_00, and scene0645\_00.
The 19 categories defined by ScanNet used for text queries are: wall, floor, cabinet, bed, chair, sofa, table, door, window, bookshelf, picture, counter, desk, curtain, refrigerator, shower curtain, toilet, sink, and bathtub. 15 categories are
without picture, refrigerator, shower curtain, bathtub; 10
categories are further without cabinet, counter, desk,
curtain, sink.

\section{Additional Experimental Results}
\subsection{Additional Qualitative Results}
Fig.~\ref{006}  presents additional qualitative results for the task of object selection in 3D space on the LERF dataset. 
Fig.~\ref{007} showcases more results of our model on the open-vocabulary 3D semantic segmentation task on the ScanNet dataset. 
These results were not included in the main manuscript due to space limitations.
Consistent with our previous observations, both OpenGaussian and InstanceGaussian exhibit limitations in handling object boundaries and in fine-grained semantic understanding. 
In contrast, our model yields results with significantly sharper and more accurate semantic interpretations.

\subsection{Additional Ablation Studies}
\label{aas}
\begin{figure*}[t]
	\centering
	\includegraphics[width=0.94\linewidth]{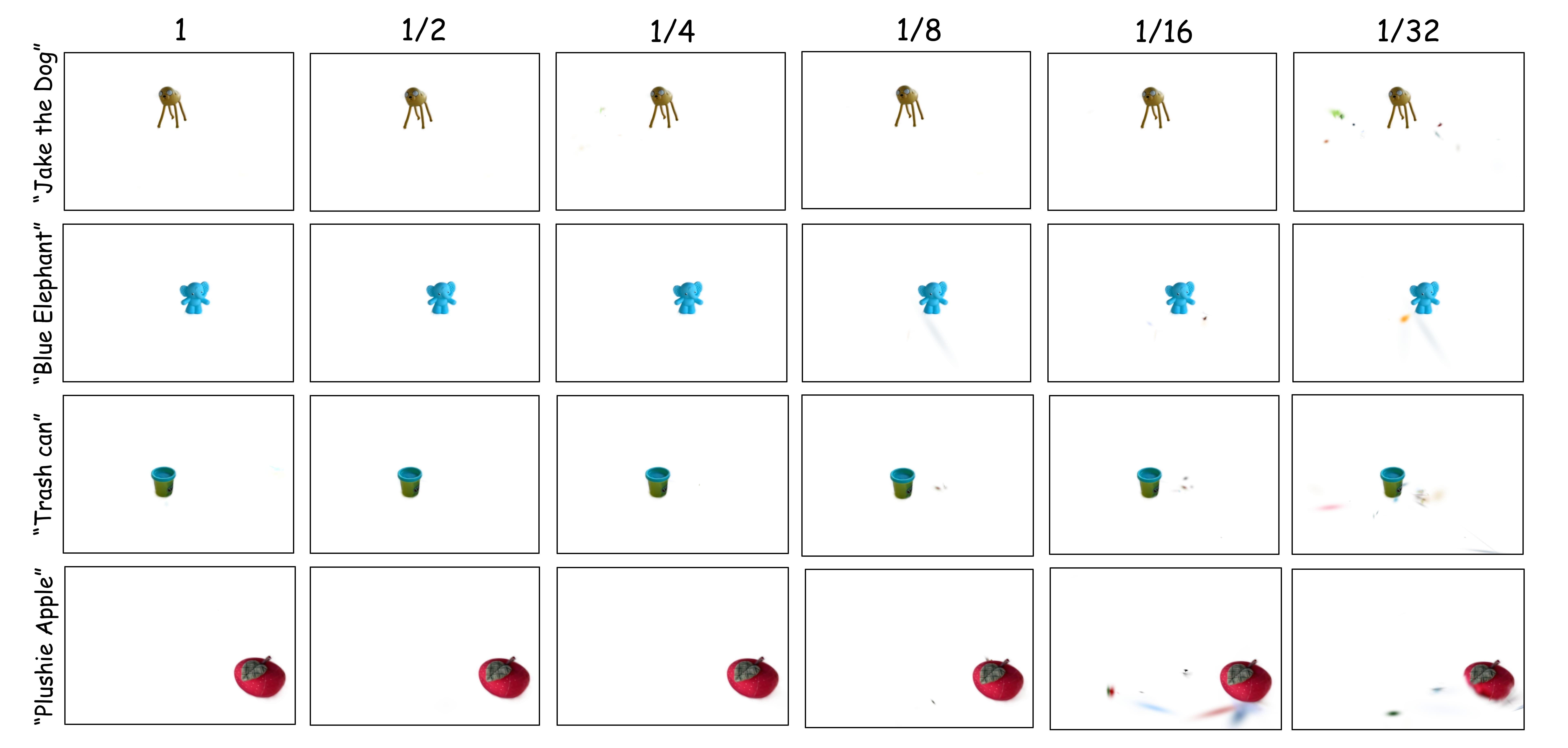}
	\caption{Open-Vocabulary 3D Object Extraction from Sparse Masks. We perform an open-vocabulary 3D object extraction task on the \texttt{figurines} scene from the LERF dataset, providing a progressively smaller subset of 2D masks as supervision. The results demonstrate that our model's accuracy experiences negligible degradation when using \textbf{$\ge 1/4$} of the total masks. With only $1/8$ of the masks, it still exhibits a strong capability to capture the object's geometry. Even in the extreme case with as few as $1/32$ of the masks, our model can still recover the object's coarse shape.}
	\label{015}
\end{figure*}

\textbf{Scene Understanding with Limited Mask Supervision.}   
Our method leverages a mask-matching mechanism for semantic understanding, a characteristic that enables it to perform 3D segmentation from only a sparse set of 2D masks. To validate this capability, we conduct experiments using progressively sparser subsets of 2D masks (corresponding to $1/2$, $1/4$, $1/8$, $1/16$, and $1/32$ of the total available views), while all other model settings are held constant. Finally, we perform an open-vocabulary 3D object extraction task and qualitatively evaluate the results.
As illustrated in Fig.~\ref{015}, our method exhibits high robustness to the number of provided masks. Even with masks from only $1/8$ of the views, it maintains high-quality segmentation. This demonstrates our model's high data efficiency and its ability to generalize from sparse supervision. However, when the number of masks becomes excessively sparse, such as at $1/16$ or $1/32$, a portion of the 3D Gaussians may not be observed by any masked camera view. This lack of supervision results in noticeable artifacts. Notably, the $1/32$ subset often corresponds to merely 5--10 foreground masks. While these extreme cases produce artifacts, the ability to generate a coherent result from such minimal data underscores our method's low reliance on dense supervision and corroborates its strong generalization capabilities.

\noindent \textbf{Ablation Study on Neutral Point Thresholds.}
On the LERF dataset, we investigate the influence of the entropy threshold $\tau_h$ and the opacity threshold $\tau_\alpha$ in our two-stage neutral point processing module. The results of this sensitivity analysis are presented in Tab.~\ref{tab:neutral_point_ablation}. The baseline configuration, which bypasses entropy-based filtering by setting $\tau_h=1.0$, achieves an mIoU of 53.0. A notable improvement is observed when $\tau_h$ is lowered to $0.9$, underscoring the efficacy of pruning points with high semantic ambiguity.
The necessity of the subsequent opacity-based filtering is also validated. With $\tau_h=0.9$, setting $\tau_\alpha=0$ removes all high-entropy points indiscriminately and degrades performance to 53.2 mIoU. This suggests that high-entropy points with high opacity are geometrically significant and should be retained. Peak performance is achieved at $(\tau_h, \tau_\alpha) = (0.9, 0.1)$. This configuration strikes a favorable trade-off between removing ambiguous transitional points and preserving geometrically salient structures. While the model demonstrates reasonable robustness to other settings, further reductions in $\tau_h$ to 0.8 or 0.5 yield diminished returns, likely due to the erroneous exclusion of valid surface points. Based on these findings, we adopt $\tau_h=0.9$ and $\tau_\alpha=0.1$ for all main experiments.

\noindent \textbf{Instance Feature Extraction.}
The core of our instance feature extraction module is a VLM that grounds textual queries to 3D visual features. The representational capacity of the VLM is therefore a critical determinant of performance. To investigate this dependency, we ablate the VLM component with three different pre-trained models on the LERF dataset: SenseNova 6.5 Pro, InternVL3-78B~\citep{chen2024internvlscalingvisionfoundation}, and Gemini 2.5 Pro~\citep{comanici2025gemini25pushingfrontier} .
The results, presented in Tab.~\ref{tab:vlm_ablation}, reveal a strong positive correlation between the representational power of the VLM and final segmentation accuracy. 
More specifically, employing VLMs known for more robust vision-language grounding consistently yields substantial gains in mIoU. This indicates that the quality of the semantic features provided by the VLM is a critical determinant of performance in this task. Therefore, the performance ceiling of our model is not static; it is set to rise in tandem with the ongoing evolution of Vision-Language Models.

We further analyze the method's sensitivity to the number of descriptive text prompts used for instance matching on the LERF dataset. As shown in Tab.~\ref{tab:prompt_ablation}, the relationship between prompt quantity and segmentation accuracy is non-monotonic. Starting from a single prompt, performance improves as the number of descriptors increases to five. This suggests that a richer set of semantic cues helps the VLM disambiguate instances, particularly for concepts too nuanced to be captured by a single term. However, increasing to 10 prompts leads to performance degradation. We hypothesize that an excessive number of prompts may introduce semantic noise or redundant information, thereby interfering with the VLM's feature matching process. Consequently, we use five descriptive prompts, as this configuration strikes a favorable balance between semantic richness and feature ambiguity.

\begin{table*}[!h]
	\centering
	\begin{minipage}[t]{0.48\linewidth}
		\centering
		\small
		\captionof{table}{Ablation on the choice of VLM.}
		\label{tab:vlm_ablation}
		\vspace{-2mm}
		\begin{tabularx}{\linewidth}{CC}
			\toprule
			Model & mIoU$\uparrow$ \\
			\midrule
			SenseNova 6.5 Pro & 47.0 \\
			InternVL3-78B     & 50.2 \\
			Gemini 2.5 Pro   & \textbf{54.3} \\
			\bottomrule
		\end{tabularx}
		\vspace{0cm} 
		\captionof{table}{Ablation on number of prompts.}
		\small
		\label{tab:prompt_ablation}
		\vspace{0mm}
		\begin{tabularx}{\linewidth}{CC}
			\toprule
			Number of Prompts & mIoU$\uparrow$ \\
			\midrule
			1  & 44.0 \\
			3  & 50.9 \\
			5  & \textbf{54.3} \\
			10 & 53.6 \\
			\bottomrule
		\end{tabularx}
	\end{minipage}
	\hfill
	\begin{minipage}[t]{0.48\linewidth}
		\centering
		\small
		\captionof{table}{Ablation on neutral point processing thresholds $\tau_h$ and $\tau_\alpha$.}
		\label{tab:neutral_point_ablation}
		\vspace{-1mm}
		\renewcommand{\arraystretch}{1.6}
		\begin{tabularx}{\linewidth}{CCC}
			\toprule
			$\tau_h$ & $\tau_\alpha$ & mIoU$\uparrow$  \\
			\midrule
			1.00 & /      & 53.0 \\
			0.99 & 0.1    & 53.8 \\
			0.90 & 0.5    & 53.8 \\
			0.90 & 0.1    & \textbf{54.3} \\
			0.90 & 0.01   & 54.2 \\
			0.90 & 0.0    & 53.2 \\
			0.80 & 0.1    & 53.8 \\
			0.50 & 0.1    & 53.1 \\
			\bottomrule
		\end{tabularx}
	\end{minipage}
\end{table*}

\subsection{Open-Vocabulary 3D Object Editing}
\begin{figure*}[h!]
	\centering
	\includegraphics[width=0.74\linewidth]{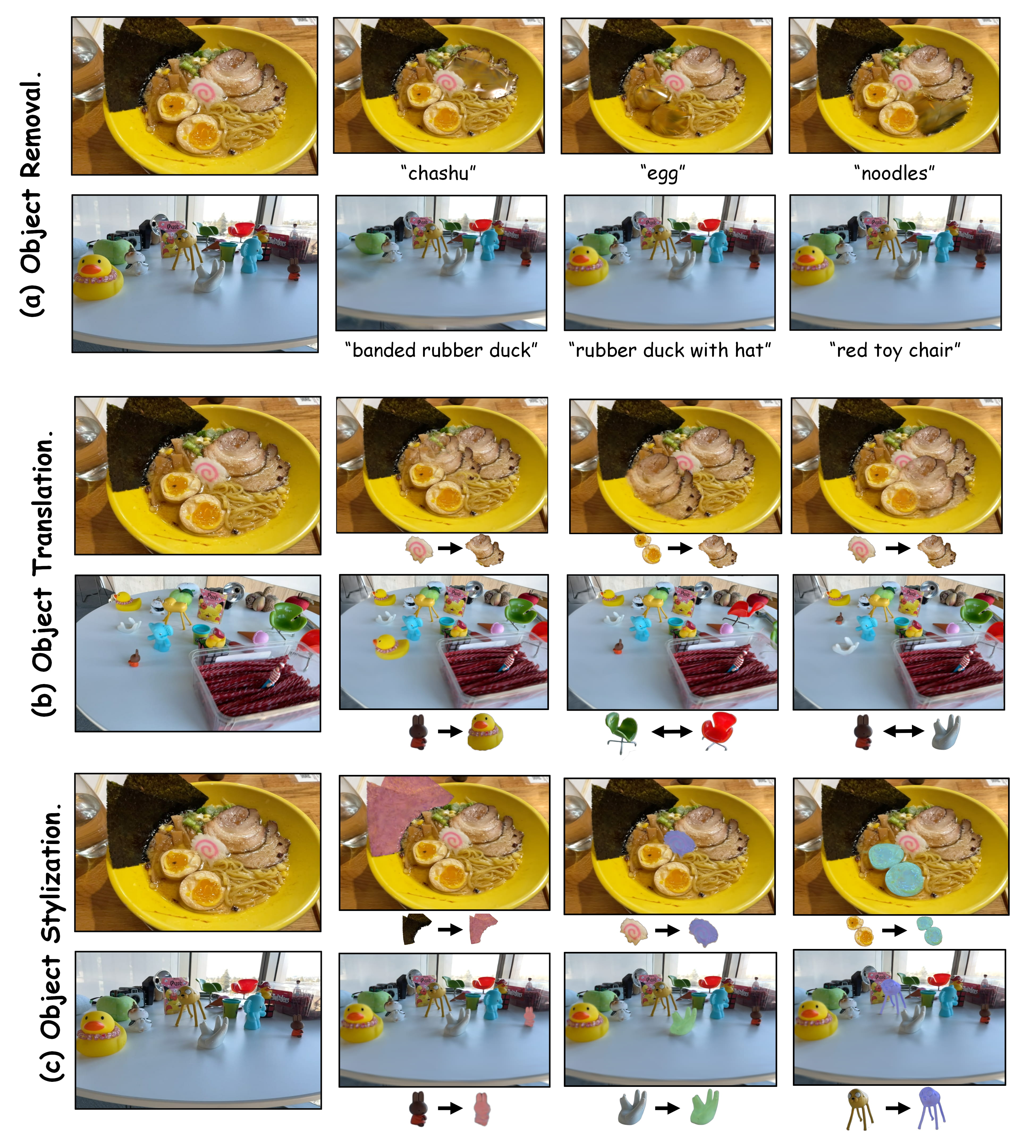}
	\caption{Demonstration of our scene editing capabilities. (a) Object Removal. (b) Object Translation. (c) Object Stylization. All manipulations are applied directly to the 3D scene rather than on the 2D rendered images.}
	\label{008}
\end{figure*}

Our method enables open-vocabulary editing of objects in 3DGS scenes by mapping a language query to corresponding instance IDs and then applying targeted manipulations.
Fig.~\ref{008} demonstrates the scene editing capabilities of our method. 
Starting from an original scene reconstructed via 3DGS, we can select an object to perform operations such as \textbf{removal} (Fig~\ref{008}(a)), \textbf{translation} (Fig.~\ref{008}(b)), or \textbf{stylization} (Fig.~\ref{008}(c)).

\subsection{Open-Vocabulary Object Extraction in Complex and Real-World Scenes}

\begin{figure*}[h!]
	\centering
	\includegraphics[width=0.7\linewidth]{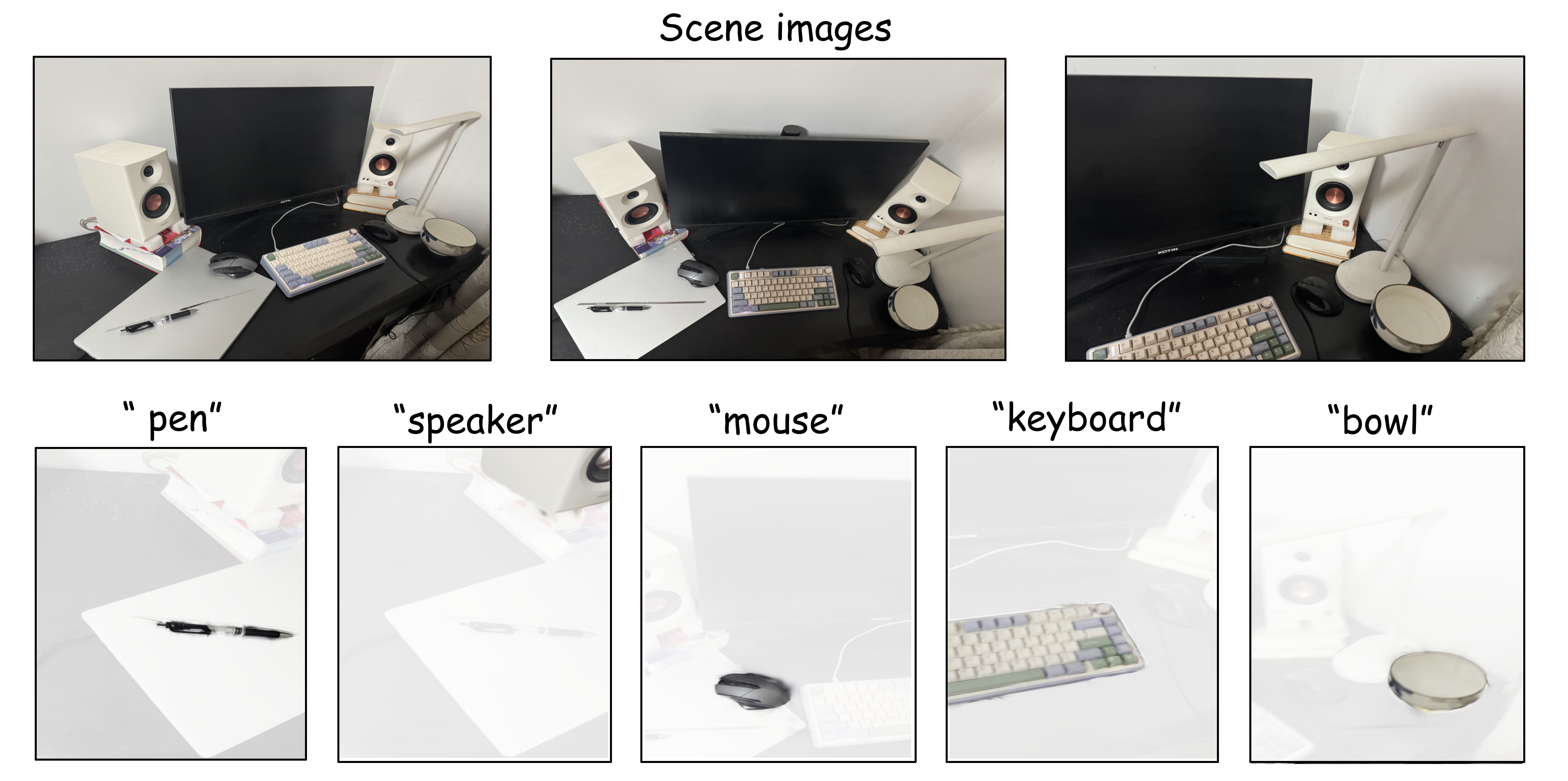}
	\caption{Qualitative results for the open-vocabulary object extraction task on a real-world scene captured with a mobile phone.}
	\label{010}
\end{figure*}

To assess its practical applicability, we validate our method on a real-world scene. We captured an office environment using a standard mobile phone and tasked our model with open-vocabulary object extraction. The qualitative results, presented in Fig.~\ref{010}, demonstrate that our model performs robustly on this in-the-wild data. This highlights the method's strong generalization capabilities and its potential for real-world applications.

\begin{figure*}[h!]
	\centering
	\includegraphics[width=1\linewidth]{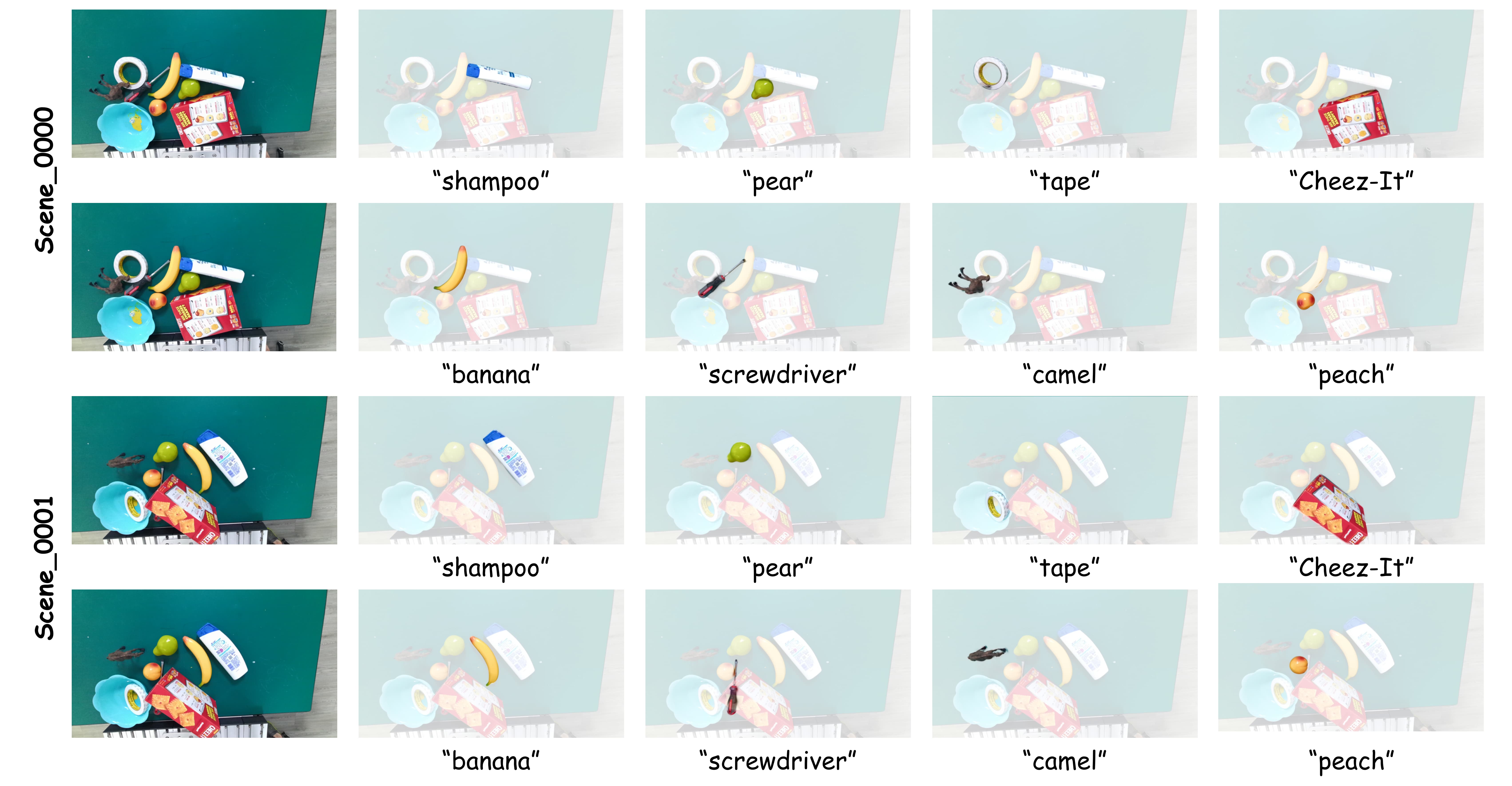}
	\caption{Qualitative results for the open-vocabulary object extraction task on the Grasp-Net dataset.}
	\label{009}
\end{figure*}

To evaluate our model's comprehension capabilities in complex scenes, we conduct experiments on the Grasp-Net dataset~\citep{fang2023robust}. This dataset is characterized by challenging object arrangements, including overlapping, adjacent, and contained instances. Despite the close proximity between instances, our model successfully distinguishes and segments them.
As shown in Fig.~\ref{009}, our method produces sharp, well-defined rendering boundaries, demonstrating its effectiveness in such challenging scenarios.
\begin{figure*}[h!]
	\centering
	\includegraphics[width=1\linewidth]{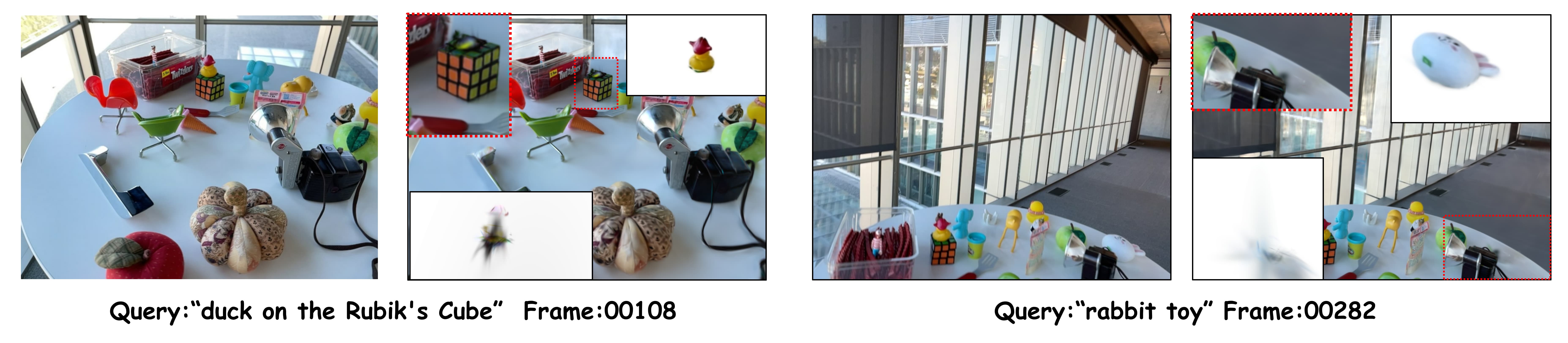}
	\caption{Qualitative results for rendering foreground, neutral, and background points on the \texttt{figurines} scene from the LERF dataset.}
	\label{016}
\end{figure*}

\section{Efficiency Analysis}
To dissect our method's efficiency, we provide a detailed component-wise runtime breakdown in Tab.~\ref{tab:runtime_full_analysis_v3}, based on the \texttt{teatime} scene in the LERF dataset, which contains 131 distinct instance categories.
The total end-to-end processing time for this complex scene is approximately 9.25 minutes (555.14s), including all computational and I/O stages. The results clearly identify the primary computational bottlenecks, with three stages accounting for over 99\% of the total computational workload: VLM Text Feature Acquisition (37.7\%), Backward Matching (32.0\%), and the initial Mask Acquisition (29.7\%). The analysis also highlights the efficiency of the neutral point processing module, which constitutes only 0.1\% of the total computational cost. This low figure indicates that the boundary refinement step is achieved with minimal performance overhead.

Notably, despite the aforementioned bottlenecks, our method's runtime holds a significant advantage over mainstream methods, which require hours of processing. For instance, in our evaluation on the LERF dataset, we found that InstanceGaussian~\citep{li_instancegaussian_2025} requires approximately 140 minutes for the 3D Gaussian training phase alone. Furthermore, our model offers potential for even greater speed. In principle, it processes each category independently, allowing for significant acceleration through parallelization. However, as a key design goal is to ensure deployability on consumer-grade hardware, this imposes a constraint on the model's total memory footprint. Consequently, we did not pursue further parallelization in the current implementation.
\begin{table}[t]
	\centering
	\footnotesize 
	\caption{Component-wise runtime breakdown for our method on the \texttt{teatime} scene in the LERF dataset. The analysis highlights that VLM inference and backward matching are the primary computational bottlenecks. All timings are in seconds, measured on a single NVIDIA Tesla V100 (32GB) GPU.}
	\label{tab:runtime_full_analysis_v3}
	\begin{tabular*}{\columnwidth}{@{\extracolsep{\fill}} l c c c}
		\toprule
		Component & Time (s) & Time / Cat. (s) & Compute \% \\
		\midrule
		\multicolumn{4}{l}{\textit{Computational Stages}} \\
		\quad Mask Acquisition & 156.99 & 1.1984 & 29.7\% \\
		\quad Backward Matching & 169.23 & 1.2919 & 32.0\% \\
		\quad Neutral Point Processing & 0.54 & 0.0041 & 0.1\% \\
		\quad Text Feature Acquisition & 199.67 & 1.5242 & 37.7\% \\
		\quad CLIP Feature Extraction & 2.63 & 0.0201 & 0.5\% \\
		\midrule
		Total Computation & 529.06 & 4.0386 & 100.0\% \\
		\midrule
		\multicolumn{4}{l}{\textit{I/O Stages}} \\
		\quad Data Loading & 16.12 & -- & -- \\
		\quad Saving Output & 9.96 & -- & -- \\
		\midrule
		Grand Total (incl. I/O) & 555.14 & -- & -- \\
		\bottomrule
	\end{tabular*}
\end{table}

\section{Analysis of Failure Cases}
\label{afc}
\begin{figure}[h!]
	\centering
	\includegraphics[width=1\linewidth]{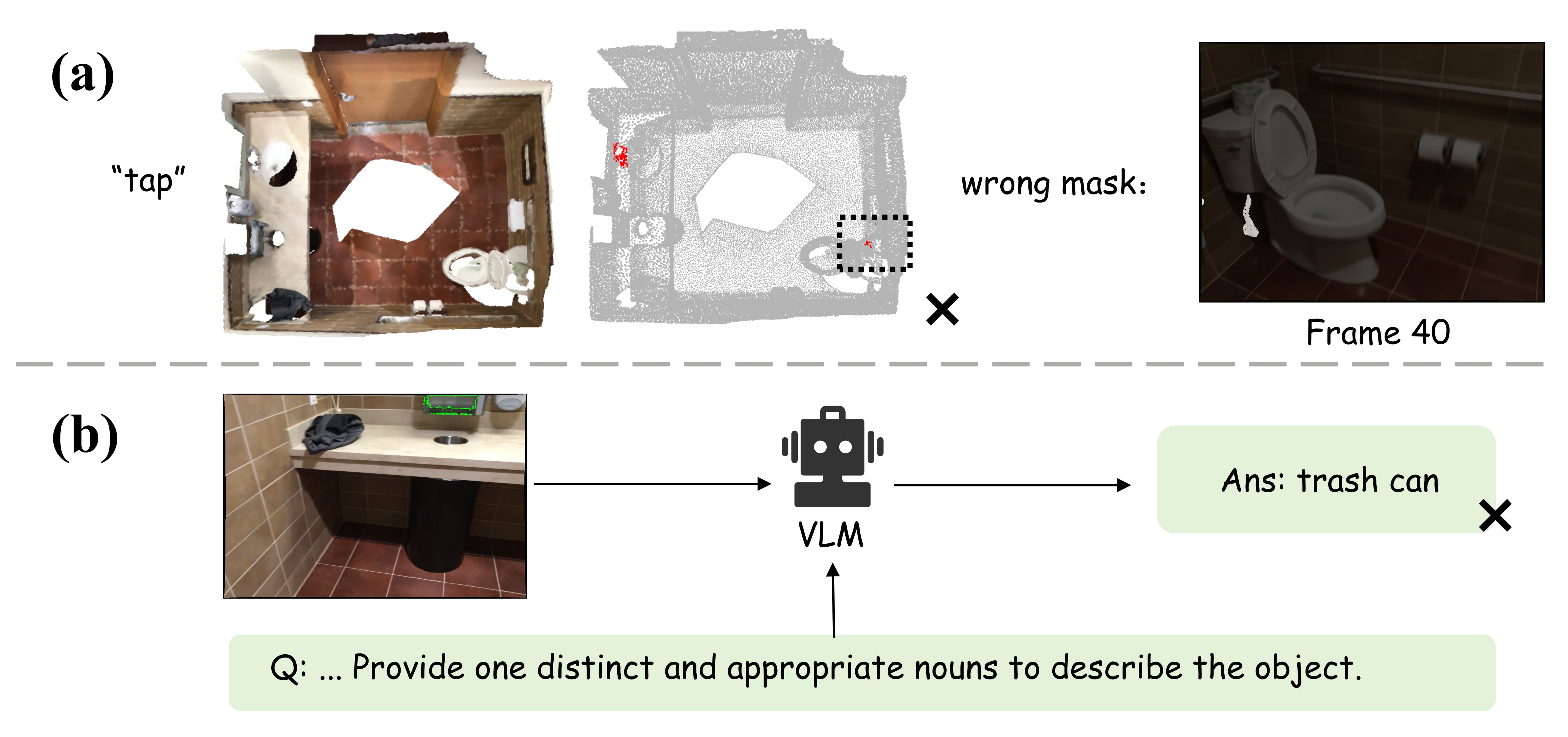}
	\caption{Examples of Failure Cases. (a) Inaccurate Masks: The segmentation model outputs incorrect 2D object masks. (b) VLM Misunderstanding: The VLM provides an incorrect object name for the given input images.}
	\label{011}
\end{figure}

\textbf{Impact of Mask Inaccuracy.}
Our method demonstrates considerable robustness to sporadic segmentation errors, provided that the initial masks generated by DAM2SAM are generally accurate. 
However, when these masks suffer from large-scale or frequent inaccuracies, our model can produce erroneous foreground-background distinctions during the backward weight accumulation process. 
This, in turn, adversely affects the final segmentation accuracy, as illustrated in a failure case in Fig.~\ref{011}(a).

\noindent \textbf{Mismatches from the VLM.}
Incorrect matching can also arise from the VLM itself, attributable to two primary sources, as shown in Fig.~\ref{011}(b).
First, ambiguous segmentation masks or challenging viewing angles in the input images can provide misleading guidance to the VLM. 
Second, inherent limitations in the VLM's comprehension capabilities can lead to incorrect judgments even with clear inputs. 
Either type of error can result in incorrect category assignments, ultimately causing the point clusters to be mismatched with the intended text query.

\section{Discussion}
\subsection{Neutral Points}

Prior work on so-called \enquote{boundary points} ~\citep{li_gradiseg_2024,zhang2025cobgs} has primarily focused on refining their positions through dedicated training strategies to enhance semantic understanding.
However, while repositioning these boundary points can improve semantic segmentation accuracy, it often compromises the realism and fidelity of the final rendering. 
This trade-off arises because boundary points include a special subset of points that belong neither to the foreground nor the background. 
These points serve as transitional elements that are crucial for ensuring rendering realism but lack specific semantic meaning. We term these as \textbf{neutral points}.

Neutral points are abundant in 3DGS scenes, making them non-negligible for semantic understanding.
Nevertheless, accurately identifying and removing these neutral points in an unsupervised manner remains a significant challenge. 
In our implementation, precisely filtering out these points during the matching stage is difficult due to computational efficiency constraints. 
In Fig.~\ref{016}, we present the visualization of neutral points from our model on the LERF dataset.
Developing more effective methods to model and eliminate neutral points is a key direction for future improvement of our method.

\subsection{Diversity of Semantic Categories}

\begin{figure}[h!]
	\centering
	\includegraphics[width=\linewidth]{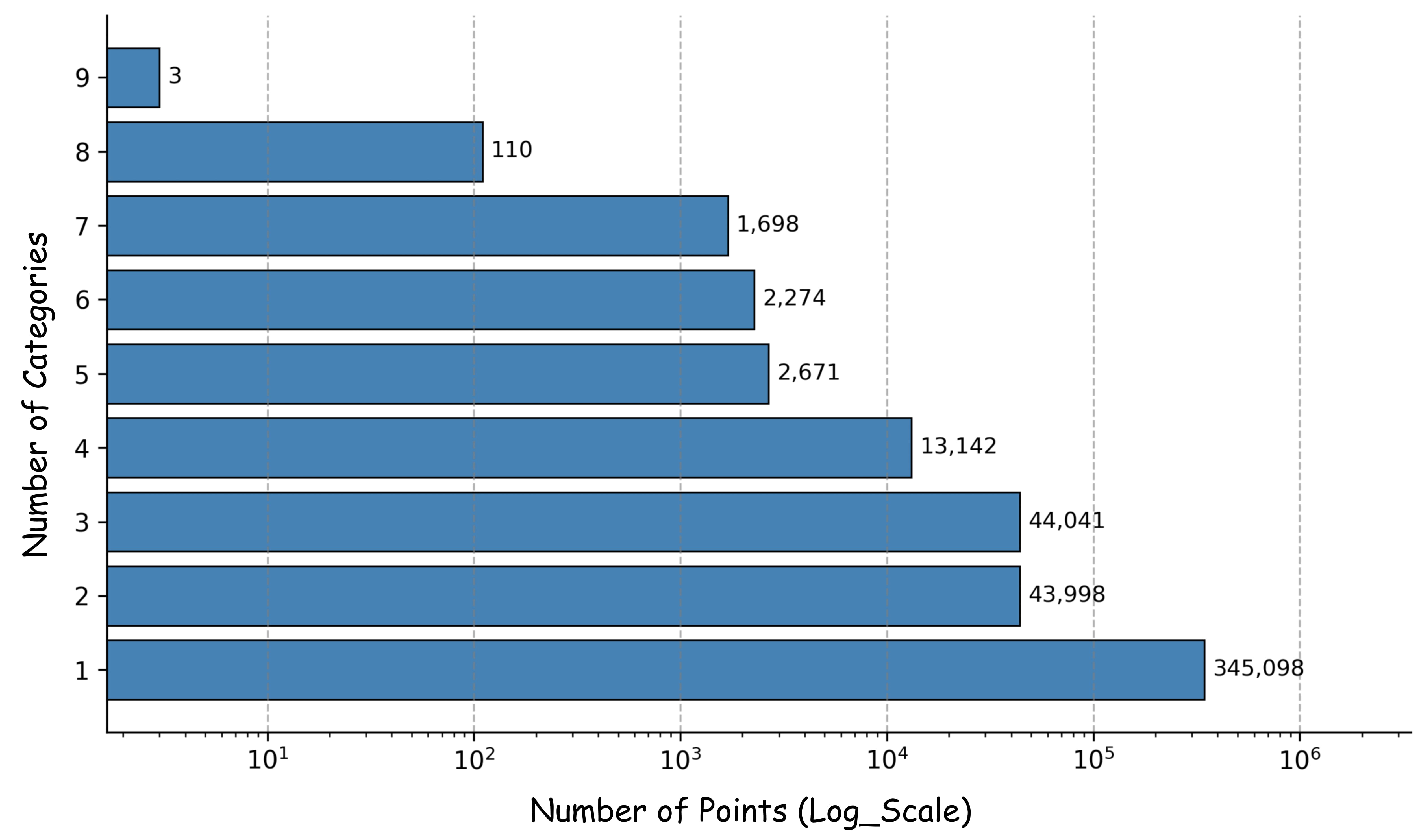}
	\caption{Category distribution of visible Gaussian points in the \texttt{teatime} scene from the LERF dataset.}
	\label{012}
\end{figure}

Prior work has noted that a single Gaussian point can belong to multiple semantic categories~\citep{shen2024flashsplat,qin_langsplat_2024,shi_language_2024}. To verify this phenomenon, we conduct a statistical analysis of the semantic categories for all 3D Gaussian points within the \texttt{teatime} scene of the LERF dataset, as illustrated in Fig.~\ref{012}. Our analysis reveals that approximately 25\% of all visible 3D points exhibit multi-dimensional semantic attributes. In the context of our model, this means a substantial portion of 3D Gaussian points inherently possess multiple semantic labels simultaneously. For instance, a single point on a tree branch may belong to the categories of \enquote{branch}, \enquote{tree}, and \enquote{vegetation} all at once.This phenomenon is consistent with how humans perceive 3D environments.

This semantic diversity suggests that relying on a single semantic label is often insufficient to comprehensively describe the properties of a point. 
Therefore, this inherent polysemy must be fully considered when performing 3D semantic understanding.

\begin{figure}[h!]
	\centering
	\includegraphics[width=0.9\linewidth]{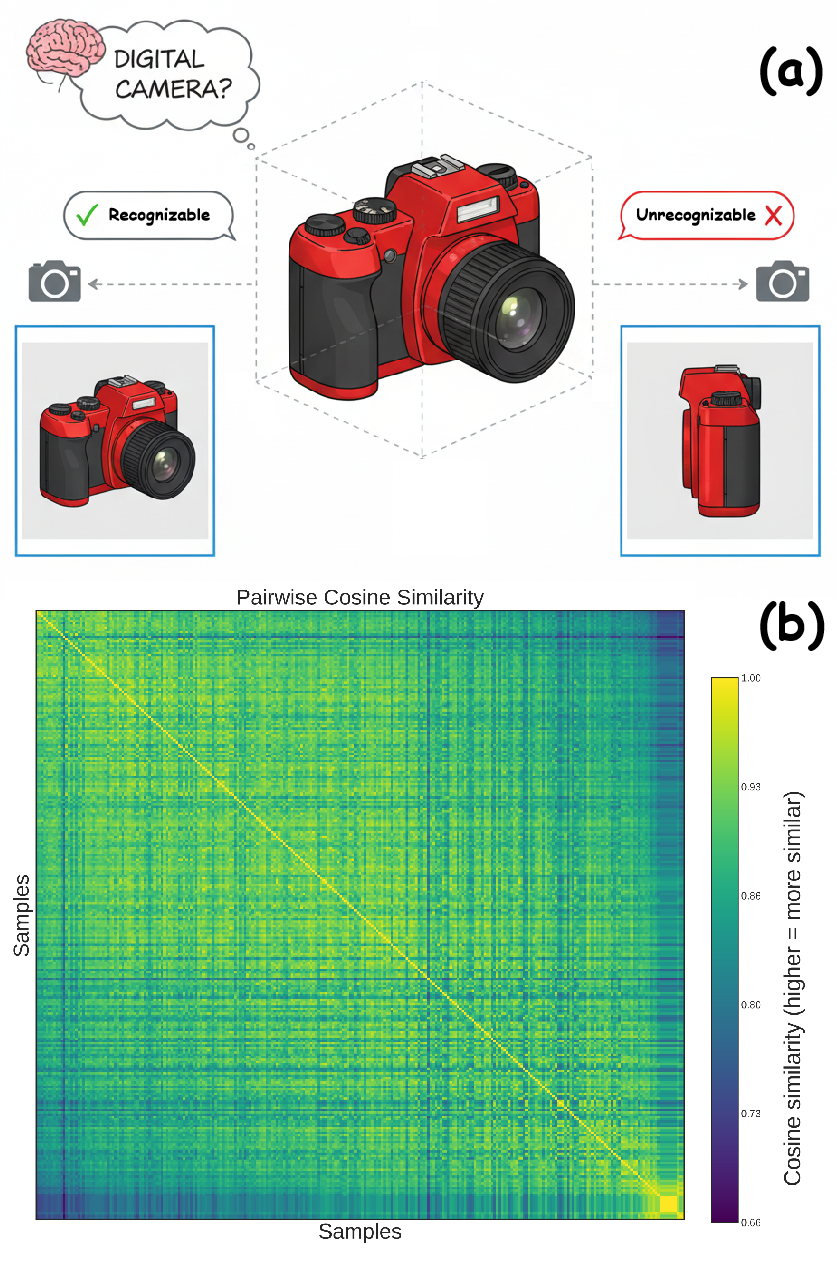}
	\caption{Illustration of the inconsistency of semantic features across different viewpoints. (a) The same object can present different semantic characteristics from different viewpoints. (b) Visualization of feature similarity for the \enquote{Jake the Dog} object in the \texttt{figurines} scene. The plot shows the cosine similarity scores between feature vectors from different views; a higher value (closer to 1) indicates that the features are more similar.}
	\label{013}
\end{figure}

\subsection{Inconsistency of Semantic Features}
\label{isf}
Our work diverges from the common practice in related literature of feeding masked object regions into a CLIP image encoder to obtain semantic features. 
This decision is based on the observation that for the same object, its semantic features can exhibit significant variations across different viewpoints~\citep{cen_tackling_2025}.
As shown in Fig.~\ref{013}(a), acquiring accurate CLIP image features becomes more challenging from certain angles. 
Due to the existence of such views, strategies like selecting the features from the view with the largest mask area or averaging the features across all views inevitably introduce errors.

To validate this phenomenon, we selected the \enquote{Jake the Dog} object from the \texttt{figurines} scene in the LERF dataset and extracted its CLIP image features from multiple viewpoints. 
A visualization of these features is presented in Fig.~\ref{013}(b). 
The figure clearly shows that even for the same object, the semantic features vary noticeably with the observation angle. 
This feature inconsistency suggests that conventional strategies based on single-mask or averaged-mask feature extraction can lead to information loss, thereby degrading matching performance. 
In contrast, our VLM-based feature extraction approach alleviates this issue to a certain extent, enhancing the stability and robustness of the semantic representation.

\end{document}